\documentclass[letterpaper,10pt,journal,twoside]{IEEEtran}

\IEEEoverridecommandlockouts %

\newcommand\copyrighttext{%
  \scriptsize \textcopyright 2023 IEEE. Personal use of this material is permitted.
  Permission from IEEE must be obtained for all other uses, in any current or future
  media, including reprinting/republishing this material for advertising or promotional
  purposes, creating new collective works, for resale or redistribution to servers or
  lists, or reuse of any copyrighted component of this work in other works.
  DOI: \href{https://doi.org/10.1109/LRA.2023.3253025}{10.1109/LRA.2023.3253025}}
\newcommand\copyrightnotice{%
\setlength{\fboxsep}{2pt}
\begin{tikzpicture}[remember picture,overlay]
\node[anchor=south,yshift=10pt] at (current page.south) {\fbox{\parbox{\dimexpr\textwidth-\fboxsep-\fboxrule\relax}{\copyrighttext}}};
\end{tikzpicture}%
}

\usepackage{amsmath} %
\usepackage{amssymb} %
\usepackage{amsfonts}
\usepackage[sort,compress]{cite}
\usepackage[usenames,dvipsnames]{xcolor}
\usepackage{graphicx}
\usepackage{tabularx}
\usepackage{multirow}
\usepackage{mathtools}
\usepackage{esvect} %
\usepackage[]{siunitx}
\usepackage{caption}
\usepackage[pdftex,pdfstartview={FitV},pdfpagelayout={TwoColumnLeft},bookmarksopen=true,plainpages=false,colorlinks=true,linkcolor=red,citecolor=green,urlcolor=black,filecolor=black ,pagebackref=false,hypertexnames=false,plainpages=false,pdfpagelabels]{hyperref}
\usepackage[T1]{fontenc} %
\usepackage{mathtools, cuted} %
\usepackage{bm}
\usepackage{xspace}
\usepackage{arydshln}

\usepackage{etoolbox}
\usepackage[normalem]{ulem}
\robustify\bfseries
\robustify\uline

\definecolor{cgreen}{RGB}{0,170,0}
\definecolor{xred}{RGB}{210,0,0}
\definecolor{vred}{RGB}{220,20,60}
\definecolor{vorange}{RGB}{210,115,0} %

\usepackage{pifont}%
\newcommand{\cmark}{{\color{cgreen}{\ding{51}}}}%
\newcommand{\xmark}{{\color{xred}{\ding{55}}}}%

\newcolumntype{H}{>{\setbox0=\hbox\bgroup}c<{\egroup}@{}}

\usepackage{enumerate}
\usepackage{balance} %
\usepackage{booktabs} %
\usepackage[export]{adjustbox} %

\newcolumntype{x}[1]{%
>{\raggedleft\hspace{0pt}}p{#1}}%

\definecolor{scan1blue}{HTML}{0072bd}
\definecolor{scan2red}{HTML}{d95319}

\usepackage{algorithm}
\usepackage[noend]{algpseudocode} %
\definecolor{commentclr}{RGB}{34, 139, 34}
\newcommand{\algcomment}[1]{{\color{commentclr}\% #1}}

\usepackage{tikz}
\usetikzlibrary{plotmarks}
\usetikzlibrary{positioning}

\usepackage{subcaption}
\DeclareCaptionLabelSeparator{periodspace}{.\quad}
\newcommand{\ra}[1]{\renewcommand{\arraystretch}{#1}}

\usepackage{amsthm}
\usepackage{thmtools, thm-restate}
\newtheorem{lemma}{Lemma}
\newtheorem{remark}{Remark}
\newtheorem{prop}{Proposition}

\theoremstyle{definition}

\makeatletter
\newcommand{\newreptheorem}[2]{%
\newtheorem*{rep@#1}{\rep@title}%
\newenvironment{rep#1}[1]{%
 \def\rep@title{#2 \ref*{##1}}%
 \begin{rep@#1}}%
 {\end{rep@#1}}}
\makeatother

\usepackage{letltxmacro}
\LetLtxMacro\orgvdots\vdots
\LetLtxMacro\orgddots\ddots

\makeatletter
\DeclareRobustCommand\vdots{%
  \mathpalette\@vdots{}%
}
\newcommand*{\@vdots}[2]{%
  \sbox0{$#1\cdotp\cdotp\cdotp\m@th$}%
  \sbox2{$#1.\m@th$}%
  \vbox{%
    \dimen@=\wd0 %
    \advance\dimen@ -3\ht2 %
    \kern.5\dimen@
    \dimen@=\wd2 %
    \advance\dimen@ -\ht2 %
    \dimen2=\wd0 %
    \advance\dimen2 -\dimen@
    \vbox to \dimen2{%
      \offinterlineskip
      \copy2 \vfill\copy2 \vfill\copy2 %
    }%
  }%
}
\DeclareRobustCommand\ddots{%
  \mathinner{%
    \mathpalette\@ddots{}%
    \mkern\thinmuskip
  }%
}
\newcommand*{\@ddots}[2]{%
  \sbox0{$#1\cdotp\cdotp\cdotp\m@th$}%
  \sbox2{$#1.\m@th$}%
  \vbox{%
    \dimen@=\wd0 %
    \advance\dimen@ -3\ht2 %
    \kern.5\dimen@
    \dimen@=\wd2 %
    \advance\dimen@ -\ht2 %
    \dimen2=\wd0 %
    \advance\dimen2 -\dimen@
    \vbox to \dimen2{%
      \offinterlineskip
      \hbox{$#1\mathpunct{.}\m@th$}%
      \vfill
      \hbox{$#1\mathpunct{\kern\wd2}\mathpunct{.}\m@th$}%
      \vfill
      \hbox{$#1\mathpunct{\kern\wd2}\mathpunct{\kern\wd2}\mathpunct{.}\m@th$}%
    }%
  }%
}
\makeatother

\let\oldnl\nl%
\newcommand{\nonl}{\renewcommand{\nl}{\let\nl\oldnl}}%
\makeatother

\def\bn{\mathbb N}

\def\br{\mathbb R}

\def\sl{\mathcal{L}}

\def\tr{\mathrm{tr}}

\newcommand\eqdef{\mathrel{\overset{\makebox[0pt]{\mbox{\normalfont\tiny def}}}{=}}}

\graphicspath{{figures/}}

\usepackage{svg}
\svgpath{{figures/}}

\newcommand{\mixeropt}{{MIXER}\textsuperscript{*}}

\newcommand{\rev}[1]{#1}

\title{MIXER: Multiattribute, Multiway Fusion of Uncertain Pairwise Affinities}

\author{Parker C. Lusk,  Kaveh Fathian, Jonathan P. How%
    \thanks{Manuscript received: October, 18, 2022; Revised January, 12, 2023; Accepted February, 12, 2023.}%
    \thanks{This paper was recommended for publication by Editor Eric Marchand upon evaluation of the Associate Editor and Reviewers' comments.
This work was supported by the Ford Motor Company and ARL DCIST under Cooperative Agreement W911NF-17-2-0181.} %
\thanks{The authors are with the Department of Aeronautics and Astronautics, Massachusetts Institute of Technology.
        {\tt\footnotesize \{plusk, kavehf, jhow\}@mit.edu}}%
}

\begin{document}

\maketitle
\copyrightnotice

\markboth{IEEE Robotics and Automation Letters. Preprint Version. Accepted February, 2023}
{Lusk \MakeLowercase{\textit{et al.}}: MIXER: Multiattribute, Multiway Fusion of Uncertain Pairwise Affinities} 

\begin{abstract} 
We present a multiway fusion algorithm capable of directly processing uncertain pairwise affinities.
In contrast to existing works that require initial pairwise \emph{associations}, our MIXER algorithm improves accuracy by leveraging the additional information provided by pairwise affinities.
Our main contribution is a multiway fusion formulation that is particularly suited to processing non-binary affinities and a novel continuous relaxation whose solutions are \emph{guaranteed} to be binary, thus avoiding the typical, but potentially problematic, solution binarization steps that may cause infeasibility.
A crucial insight of our formulation is that it allows for three modes of association, ranging from non-match, undecided, and match.
Exploiting this insight allows fusion to be delayed for some data pairs until more information is available, which is an effective feature for fusion of data with multiple attributes/information sources.
We evaluate MIXER on typical synthetic data and benchmark datasets and show increased accuracy against the state of the art in multiway matching, especially in noisy regimes with low observation redundancy.
Additionally, we collect RGB data of cars in a parking lot to demonstrate MIXER's ability to fuse data having multiple attributes (color, visual appearance, and bounding box).
On this challenging dataset, MIXER achieves $74\%$ $\mathrm{F}_1$ accuracy and is $49$x faster than the next best algorithm, which has $42\%$ accuracy.
\rev{Open source code is available at \href{https://github.com/mit-acl/mixer}{https://github.com/mit-acl/mixer}.}
\end{abstract}

\begin{IEEEkeywords}
Sensor Fusion; Sensor Networks; Mapping
\end{IEEEkeywords}

\section{Introduction}\label{sec:intro}

\IEEEPARstart{E}{stablishing} correspondences among data is a fundamental step in many computer vision, robotic perception, and estimation tasks.
For example, in structure from motion, SLAM, and object tracking, feature/object matches between sensor observations are desired so that motion can be estimated and landmark maps can be built.
Often, pairwise correspondences are identified by solving a linear assignment problem (LAP) after constructing a matrix of feature similarities.
In practice, observations are contaminated with noise and false detections that can cause the LAP, typically solved with the Hungarian~\cite{kuhn1955hungarian} algorithm, to produce incorrect matches (outliers).
When fusing observations from different views or times (e.g., map fusion in SLAM~\cite{aragues2011consistent}), these outlier matches may cause distinct data to be combined, resulting in inconsistencies~\cite{fathian2020clear}.

Multiway matching attempts to correct outliers by exploiting observation redundancy and enforcing \textit{cycle consistency}---a property stating that the composition of pairwise matchings over cycles must be identity.
Many state-of-the-art multiway matching algorithms do this by permutation synchronization~\cite{Pachauri2013,chen2014near,Maset2017,leonardos2017distributed,bernard2019synchronisation,leonardos2018distributed,leonardos2020low,fathian2020clear,shi2021fcc,li2022fast} of pairwise correspondences, which are improved and made consistent by joint optimization.
These methods are effective at multiway matching when binary matchings are available; however, their use of (partial) permutation matrices is akin to late fusion~\cite{snoek2005early,gadzicki2020early} and precludes them from using all available information, i.e., multiway matching is performed \emph{after} each pairwise affinity matrix is pre-processed to create binary pairwise matches.

\begin{figure}[t]
    \centering
    \includegraphics[trim=0 2cm 0 0,clip,width=\columnwidth]{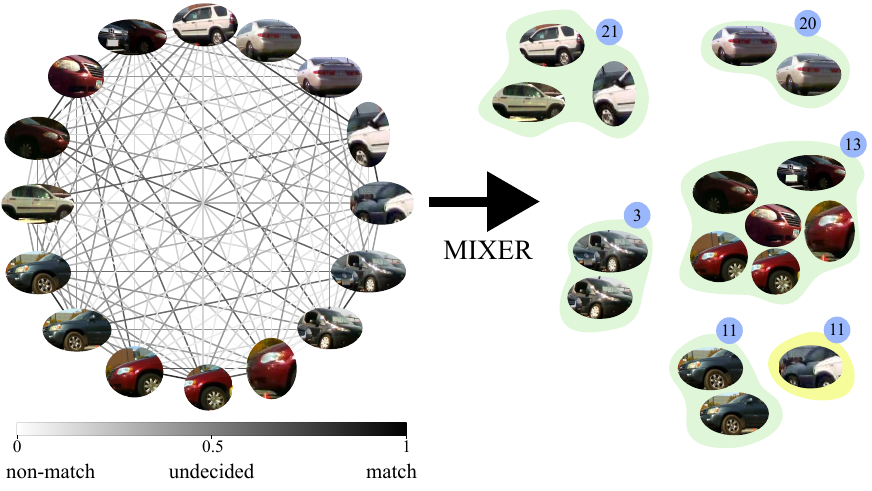}
    \caption{
    Multiattribute, multiway fusion example of car detections from different images.
    Each detection has SIFT, color, and bounding box attributes used to perform pairwise affinity scoring.
    Pairwise affinities are frequently uncertain, due to noisy detections and scoring processes, and the resulting multiway affinity matrix (graph representation, left) is inconsistent.
    Leveraging cycle consistency, distinctness constraints, and three modes of association, MIXER achieves high accuracy data fusion, correctly clustering identical cars (green clusters, right).
    If observations are too uncertain, MIXER tends to sacrifice recall in favor of precision, illustrated in the case of car 11---the detection in the yellow cluster is too ambiguous to be fused.
    }
    \label{fig:teaser-image}
\end{figure}

In contrast, we present MIXER (Multiway affinity matrIX fusER), an algorithm that produces cycle consistent multiway matches directly from noisy and uncertain pairwise affinities, i.e., in an early fusion sense.
Similar to the conclusions of data fusion works in other contexts~\cite{snoek2005early,gadzicki2020early}, our experiments show that MIXER's early multiway fusion can yield a significant performance increase over existing late fusion approaches.
Specifically, direct access to affinities enables a key property of our optimization formulation of multiway fusion.
Because of our squared Frobenius objective, our formulation gives rise to three modes of association: non-match, uncertain match, and match, corresponding to $0$, $0.5$, and $1$ affinities, respectively.
These three modes also appear in a generalization of pairwise LAP called maximum-weight matching (MWM)~\cite{ramshaw2012minimum,duan2014linear}, where affinities less than or equal to a threshold (i.e., $0.5$) will never be associated because they penalize the objective.
MIXER extends these ideas to the multiway case (see Remark~\ref{rem:multiway-mwm}), balancing the three association modes in conjunction with cycle consistency and other problem constraints to achieve high accuracy in the presence of uncertain pairwise affinity scores.
An example of MIXER fusing together car observations is shown in Fig.~\ref{fig:teaser-image}.

The availability of three association modes is especially important when working with data having multiple modalities or attributes (e.g., color, lidar reflectance, position, bag-of-words vector~\cite{galvez2012bags}, SIFT descriptors~\cite{lowe2004sift}, bounding box~\cite{bewly2016sort}, shape~\cite{cootes1995active}, reID features~\cite{karanam2019reid}, etc.).
Because attributes may produce affinity scores in contention (e.g., an object viewed from different viewpoints may have the same color, but not have bounding box overlap), the combined affinity becomes more uncertain, trending toward $0.5$.
This allows MIXER to defer to other pairwise affinities, and to the problem constraints, before making a decision.

We formulate the early multiway fusion problem as a mixed-integer quadratic program (MIQP).
Since MIQPs are generally not scalable, we propose a novel continuous relaxation leading to approximate solutions.
The main contribution of the MIXER algorithm over similar techniques is that its solutions are \emph{guaranteed} to converge to feasible (binary) solutions of the original MIQP.
Thus, rounding/projecting results to binary values---which is required when using other techniques and may lead to infeasible solutions---is avoided.
To solve the relaxed problem efficiently, we present a projected gradient descent algorithm with backtracking line search.
This polynomial-time algorithm has worst case cubic complexity in problem size (from matrix-vector multiplies) at each iteration, and is guaranteed to converge to second-order stationary points~\cite[Prop.~7]{mokhtari2018escaping}.

We evaluate MIXER on synthetic and real-world datasets and compare the results with state-of-the-art multiway data association algorithms.
Our synthetic analysis shows an empirically tight optimality gap of MIXER solutions with respect to the global minimum of the MIQP, while achieving an average runtime of \SI{8}{\milli\second}---four orders of magnitude faster than solving the MIQP with a general-purpose solver.
Our real-world evaluation considers nine multiway matching benchmark datasets, showing that even in the \emph{single} attribute early fusion setting, MIXER is able to achieve high accuracy, superior to the state of the art.
Finally, we collect our own dataset of RGB images recorded in a parking lot and attempt to fuse observations of cars seen from multiple viewpoints.
Three complementary attributes of cars are extracted (bounding box, color, and SIFT visual appearance) and we show that MIXER significantly outperforms existing algorithms, increasing $\mathrm{F}_1$ score over the next-best result by $32\%$ while being $49$x faster.
In summary, our main contributions are:
\begin{itemize}
\itemsep0em
    \item A principled formulation of multiway fusion as an MIQP that when approached in an early fusion framework leads to three states of association and a multiway extension of the pairwise MWM problem.
    \item A novel continuous relaxation of the multiway fusion MIQP leading to MIXER, a polynomial-time algorithm based on projected gradient descent that converges to stationary points.
    \item Theoretical analysis of the continuous relaxation, showing that the MIXER algorithm is guaranteed to converge to feasible, binary solutions of the original MIQP.
    \item Substantial accuracy and timing improvements over state-of-the-art multiway maching algorithms on standard benchmarks and in a challenging, self-collected RGB dataset with three distinct attributes.
\end{itemize}

\subsection{Related Work}\label{sec:related_work}

\textbf{Pairwise association.}
Associating elements from two sets based on inter-element similarity scores is traditionally formulated as a LAP~\cite{burkard2009assignment}, which can be optimally solved in polynomial time using, e.g., the Hungarian~\cite{kuhn1955hungarian} algorithm.
The Hungarian algorithm produces \emph{perfect matchings} (one-to-one correspondence with all elements matched) for \emph{balanced} matching problems (same number of elements in both sets).
Imperfect (one-to-one correspondence, all items need not be matched) or unbalanced matching problems can be reduced to perfect, balanced matching to be solved with Hungarian~\cite{ramshaw2012minimum}.
The case where an imperfect matching of any size is sought that maximizes the possible benefit is called the maximum-weight matching (MWM) problem~\cite{duan2014linear}.

If elements have underlying structure (e.g., geometric structure of 3D point clouds) that should be included in the association decision, the problem can be formulated as a quadratic assignment problem (QAP)~\cite{lawler1963quadratic} (equivalently, graph matching~\cite{conte2004thirty}).
Unlike the LAP, the QAP (and its equivalent graph matching formulation) is, in general, NP-hard~\cite{sahni1976p}.
Generalizations of graph matching introduce additional clique constraints to improve robustness~\cite{lusk2021clipper}.

\textbf{Multiway association.}
Multiway data association frameworks jointly associate elements across multiple sets to ensure cycle consistency of associations.
Multiway association can be formulated as a permutation synchronization problem~\cite{Pachauri2013}, which is computationally challenging due to its binary domain. 
With the exception of expensive combinatorial methods~\cite{Zach2010,Nguyen2011}, existing works consider relaxations to obtain approximate answers. 
\rev{These approximations include 
spectral relaxation \cite{Pachauri2013, Maset2017},
convex relaxation \cite{chen2014near, Hu2018, Yu2016, leonardos2017distributed, leonardos2018distributed, de2021efficient},
matrix factorization \cite{zhou2015multi, bernard2019synchronisation, leonardos2020low},
filtering by cluster-consistency statistics~\cite{shi2021fcc},
message passing~\cite{li2022fast},
iteratively reweighted least squares~\cite{shi2020robust}, %
and graph clustering \cite{yan2015multi, tron2017quickmatch, serlin2020distributed, fathian2020clear}.}
With the exception of \cite{tron2017quickmatch,zhou2015multi}, most of the aforementioned methods were originally
designed with permutation synchronization in mind and test only with binary associations as input, i.e., in a late fusion approach.
Additionally, while our MIQP formulation has been used before~\cite{leonardos2018distributed,leonardos2020low}, our particular relaxation allows us to guarantee that MIXER will converge to feasible, binary solutions, avoiding the final rounding procedure necessary in other works.
\rev{Lastly, when underlying structure is incorporated, the formulation becomes a multi-graph matching problem \cite{yan2013joint, shi2016tensor, yan2015multi, Swoboda2019, dupe2022kernelized} (equivalently, multi-QAP), which is considerably more computationally challenging.}

\section{Problem Formulation}\label{sec:formulation}

\begin{figure}[t]
    \centering
    \includegraphics[trim=0 15.25mm 0 0,clip,width=1.0\columnwidth]{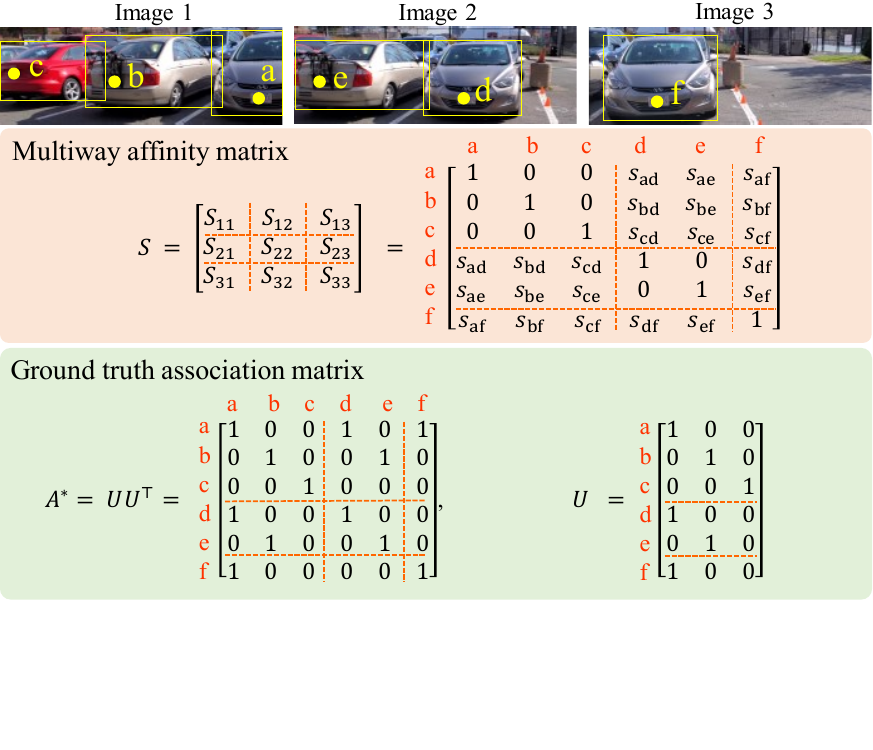}
    \caption{ 
    Example with $k=3$ cars across $n=3$ images, with a total of $m=6$ observations.
    Bounding box overlap (had they been drawn on the same image) gives the similarity score between two cars.
    The multiway affinity matrix $S$ and its corresponding ground truth $A^*$ are shown.
    }
  \label{fig:example}
\end{figure}

In this section, we formalize a principled framework to solve the multiway fusion problem.
Consider $n$ sets of data $\mathcal{S}_i,\,i=1,\dots,n$, with cardinality $|\mathcal{S}_i|=m_i$ and let $m=\sum_{i=1}^nm_i$.
We define the \emph{universe} as $\mathcal{U}\eqdef\cup_i\,\mathcal{S}_i$, with $|\mathcal{U}|=k\leq m$ distinct elements across all sets.
For example, Fig.~\ref{fig:example} shows $n=3$ images, each with $m_i$ car detections with bounding box attributes.
These are denoted by $\mathcal{S}_1 \eqdef \{ a, b, c \}$,  $\mathcal{S}_2 \eqdef \{d, e \}$, and  $\mathcal{S}_3 \eqdef \{f\}$. 
Assuming that we know observations $a,d,f$ and $b,e$ represent the same cars, the universe is $\mathcal{U} \eqdef \{a, b, c\}$ and has $k=3$ elements.

Given two sets $\mathcal{S}_i$ and $\mathcal{S}_j$, we define the \textit{pairwise affinity matrix} between observations as
\begin{equation} \label{eq:pairwise-affinity}
S_{ij} \eqdef
\begin{bmatrix}
s_{11} & \cdots & s_{1m_j} \\
\vdots & \ddots & \vdots \\
s_{m_i1} & \cdots & s_{m_im_j}
\end{bmatrix}
\in[0,1]^{m_i\times m_j},
\end{equation}
where $s_{ab}\in[0,1]$ denotes the similarity between elements $a \in \mathcal{S}_i$ and $b \in \mathcal{S}_j$.
Scores of $0$, $0.5$, and $1$ correspond to maximum dissimilarity, maximum uncertainty/no preference, and maximum similarity, respectively.
The \textit{multiway affinity matrix} between all sets is defined as the symmetric matrix
\begin{equation} \label{eq:multiway-affinity}
S \eqdef
\begin{bmatrix}
    S_{11} & \cdots & S_{1n} \\
    \vdots  & \ddots & \vdots \\
    S_{n1} & \cdots & S_{nn} 
\end{bmatrix}
\in[0,1]^{m\times m },
\end{equation}
where, by definition, $S_{ij} = S_{ji}^\top$.
The example in Fig.~\ref{fig:example} shows the multiway affinity matrix $S$ (henceforth called the affinity matrix for simplicity), where $m=6$.
Pairwise association matrix blocks are separated by dashed lines.

\textbf{Ground-truth association.}
The ground-truth pairwise affinity matrices take values of $1$ for identical objects and $0$ otherwise.
When affinity matrices are binary, like in the ground truth case, we refer to them as association matrices.
Furthermore, the ground-truth multiway association matrix, denoted by $A^*$, can be factorized as ${A^* = UU^\top}$, with the \textit{universal association matrix} $U$ is defined as
\begin{gather} \label{eq:universal-association}
U \eqdef
\begin{bmatrix} U_{1} & \cdots & U_{n} \end{bmatrix}
\in\{0,1\}^{m\times k}.
\end{gather}
Matrices $U_i\in\{0,1\}^{m_i\times k}$ represent associations between elements of sets $S_i$ and $\mathcal{U}$.
Fig.~\ref{fig:example} shows $A^*$ and $U$ for the corresponding example, with $k=3$ unique cars across all images and where matrices $U_i$ are separated by dashed lines.

\textbf{Constraints.}
Often, data association algorithms must meet certain constraints imposed by the high-level task.
The \textit{one-to-one} constraint states that an object can be associated with at most one other object. 
This constraint is satisfied if each row of $U$ has a single $1$ entry.
The \textit{distinctness} constraint states that objects within a set are distinct and therefore should not be associated.
This is satisfied if there is at most a single $1$ entry in each column of $U_i$.
When the association problem is solved across more than two sets, associations must be \textit{cycle consistent}, that is, if $a\sim b$, and $b\sim c$, then $a\sim c$.
This condition is satisfied if the association matrix can be factorized as $A=UU^\top$~\cite{zhou2015multi}.
These three constraints are crucial for detecting and correcting erroneous similarity scores and associations, but increase the difficulty and complexity of the data association procedure.

\begin{table*}[t]
\scriptsize
\centering
\caption{
Comparison of our MIXER formulation with multiway matching algorithms that also relax combinatorial problems with a Frobenius objective.
The resulting relaxations are similar, but MIXER guarantees that solutions are cycle consistent (cyc.), distinct (dis.), and binary (bin.).
Further, MIXER obtains solution using an efficient projected gradient descent (PGD) method where the projection $\Pi_\mathcal{C}$ onto the constraint set $\mathcal{C}$ can be evaluated in closed form.
When a binary solution is not guaranteed, superscripts `c', `t', and `h' indicate rounding via clustering, thresholding, and Hungarian, respectively.
Parameters $\lambda$ and $\alpha$ arise due to sparsity regularization (different from our perspective) and are set to the defaults suggested by the respective authors.
}
\setlength{\tabcolsep}{3.2pt}
\ra{1.1}
\begin{tabular}{l l c c c c c c}
\toprule
Algorithm & Objective & Constraint Set & Solution Method & $\Pi_\mathcal{C}(\cdot)$ & \multicolumn{3}{c}{Soln. Guarantees} \\
\cmidrule{6-8}
&&&&& cyc. & dis. & bin. \\
\toprule
MatchLift~\cite{chen2014near}
    & $\underset{A \in \br_+^{m\times m}}{\text{minimize}} \quad \big < A, \mathbf{1}-\tfrac{1}{\lambda}S \big >$
    & 
\parbox{0cm}{
\begin{align*}
    A_{ii}=I_{m_i},\,\forall_i,\; \begin{bsmallmatrix} k  & ~~\mathbf{1}^\top \\ \mathbf{1} & A\end{bsmallmatrix} \succeq \mathbf{0}
\end{align*}}
    & SDP/ADMM & -- & \cmark & \cmark & \xmark\,\textsuperscript{c}
    \\[-1em]
MatchALS~\cite{zhou2015multi}
    & $\underset{A\in\mathbb{R}_+^{m\times m}}{\text{minimize}} \quad \big < A, \mathbf{1}-\tfrac{1}{\alpha}S \big > + \lambda\|A\|_*$
    &
\parbox{0cm}{
\begin{align*}
    A_{ii}=I_{m_i},\,\forall_i,\; A_{ij}=A_{ji}^\top,\,\forall_{i\neq j} \\
    \mathbf{0}\leq A_{ij}\mathbf{1}\leq\mathbf{1},\; \mathbf{0}\leq A_{ij}^\top\mathbf{1}\leq\mathbf{1}
\end{align*}}
    & ADMM & LP & \xmark & \cmark & \xmark\,\textsuperscript{t}
    \\[-1em]
MatchDGD~\cite{leonardos2018distributed}
    & $\underset{U\in\mathbb{R}_+^{m\times k}}{\text{minimize}} \quad \big < UU^\top, \mathbf{1}-2S \big > + \sum_i\|I-U_iU_i^\top\|^2_F$
    &
\parbox{0cm}{
\begin{align*}
    U\mathbf{1} = \mathbf{1},\; U_i^\top\mathbf{1}\leq\mathbf{1},\,\forall_i
\end{align*}}
    & PGD & ADMM & \cmark & \cmark & \xmark\,\textsuperscript{h}
    \\[-0.5em]
MatchRTR~\cite{leonardos2020low}
    & $\underset{U\in\mathbb{R}_+^{m\times k}}{\text{minimize}} \quad \big < UU^\top, \mathbf{1}-2S \big > + \sum_i\|I-U_iU_i^\top\|^2_F$
    &
\parbox{0cm}{
\begin{align*}
    U\mathbf{1} = \mathbf{1}
\end{align*}}
    & \begin{tabular}{@{}c@{}}Riemannian \\ trust-region\end{tabular} & closed form & \cmark & \xmark & \xmark\,\textsuperscript{h}
    \\[-0em]
\midrule
MIXER
    & $\underset{U\in\mathbb{R}_+^{m\times m}}{\text{minimize}} \quad \big < U U^\top,  \mathbf{1}-2S \big > + d\big(\phi_\mathrm{orth}(U) + \phi_\mathrm{dist}(U)\big)$
    &
\parbox{0cm}{
\begin{align*}
    U\mathbf{1} = \mathbf{1}
\end{align*}
}
    & PGD & closed form & \cmark & \cmark & \cmark \\
\bottomrule
\end{tabular}
\label{tbl:relaxations}
\vskip-0.25in
\end{table*}

\textbf{Optimization problem.}
Given noisy and potentially incorrect similarity scores, our goal is to rectify their values to binary scores that respect the one-to-one, distinctness, and cycle consistency constraints.     
This goal can be formally stated as finding $U$ that solves the MIQP
\begin{gather} \label{eq:miqp}
\begin{aligned}
& \underset{U \in \{0, 1\}^{m\times k}}{\text{minimize}} & &  \big\| U\, U^\top -  S \big\|_{F}^2 \qquad  {\color{gray} \text{(cycle consistency)}}  \\
& ~~\text{subject to}  & &  U \, \mathbf{1}_m  = \mathbf{1}_m  \qquad  \quad \, {\color{gray} \text{(one-to-one)}} \\
&&&  U_i^\top \mathbf{1}_{m_i} \leq \mathbf{1}_{m_i} \qquad  {\color{gray} \text{(distinctness)}}
\end{aligned} 
\end{gather}
where $\|\cdot \|_F$ is the matrix Frobenius norm and $\mathbf{1}$ denotes the vector of all ones (with size shown as subscript).
When the universe size $k$ is known (or a good estimate $\hat{k}$ is available), this can be used directly in \eqref{eq:miqp} to determine the dimensions of $U$.
However, $k$ is rarely known in practice and can be difficult to reliably estimate, so instead we allow the algorithm to simultaneously estimate $U$ and $k$ by setting $k=m$.
Then, a solution $U$ of \eqref{eq:miqp} will have $\hat{k}$ nonzero columns, giving an estimate of the universe size.

\begin{remark}[Multiway MWM] \label{rem:multiway-mwm}
For $n=2$, problem~\eqref{eq:miqp} reduces to a LAP.
Replacing the affinity matrix by ${S\leftarrow\frac{1}{2}(\mathbf{1}-S)}$, problem~\eqref{eq:miqp} returns a perfect matching and solutions are equivalent to Hungarian solutions.
With $S$ as specified in Sec.~\ref{sec:formulation}, \eqref{eq:miqp} solves the MWM problem~\cite{duan2014linear} and can return imperfect matchings.
This generalization of the LAP is often more natural for data association since not all data should always be fused.

The traditional two-way MWM problem is often reduced so that a perfect matching exists~\cite{ramshaw2012minimum} and is solved using the Hungarian algorithm.
This is done by setting any affinity below a threshold to $0$ and then discarding any match with $0$ affinity in post-processing.
Instead, \eqref{eq:miqp} does not require this explicit thresholding and extends to $n>2$, solving the multiway maximum-weight matching problem by allowing the information from multiple pairs of views to inform the final decision.
\end{remark}

\section{Continuous Relaxation and Algorithm} \label{sec:theory}

Due to its binary domain, solving \eqref{eq:miqp} to global optimality requires combinatorial techniques that quickly become intractable as the problem size grows.
To increase scalability, the standard workaround is to relax the domain of the problem to the positive orthant. %
However, these solutions must be subsequently \textit{rounded} (i.e., projected back to binary values), which can be problematic since this may produce infeasible solutions that violate the original constraints.
A key novelty of our relaxation approach and algorithm is that solutions are \emph{guaranteed} to converge to \emph{feasible}, \emph{binary} solutions of the original problem~\eqref{eq:miqp}, thereby obviating the potentially problematic rounding step.
Towards our relaxation and guarantees, we state the following equivalent formulation of \eqref{eq:miqp}
\begin{gather} \label{eq:penalty-form}
\begin{aligned}
& \underset{U\in\{0,1\}^{m\times m}}{\text{minimize}} & &  \big < UU^\top,  \mathbf{1}-2S \big > & {\color{gray} \text{(cycle consistency)}} \\
& \text{subject to}
 &&  U\mathbf{1}_m=\mathbf{1}_m  &  {\color{gray} \text{(one-to-one constraint)}} \\
&&&  \phi_\mathrm{orth}(U) = 0  &  {\color{gray} \text{(orthogonality constraint)}} \\
&&&  \phi_\mathrm{dist}(U) = 0 & {\color{gray} \text{(distinctness constraint)}}
\end{aligned}
\end{gather}
where $\phi_\mathrm{orth}$ and $\phi_\mathrm{dist}$ are the penalty functions
\begin{equation} \label{eq:penalties}
\phi_\mathrm{orth} \eqdef  \big < U^\top U, P_o \big >, \qquad \phi_\mathrm{dist} \eqdef \big < U U^\top, P_d \big >,
\end{equation}
with penalty matrices defined as $P_o \eqdef \mathbf{1} - I$ and $P_d \eqdef \mathrm{blockdiag}(P_{d1}, \dots, P_{dn})$, and ${P_{di} \eqdef \mathbf{1}_{m_i \times m_i} - I_{m_i \times m_i}}$.
\rev{The orthogonality constraint is implicit in equation~\eqref{eq:miqp} due to the binary domain.
Both $\phi_\mathrm{orth}$ and $\phi_\mathrm{dist}$ are non-negative functions and are equal to zero if and only if the respective constraint is satisfied.
Further details on the design of the penalty functions and on the equivalence between problem~\eqref{eq:penalty-form} and problem~\eqref{eq:miqp} can be found in
Appendix~\ref{apdx:equiv-penalty-form}.}

\textbf{Relaxed Formulation.}
\rev{We relax problem~\eqref{eq:penalty-form} by removing the binary constraint and including the orthogonality and distinctness constraints into the objective, scaled by $d\ge0$,}
\begin{gather} \label{eq:relaxed}
\begin{aligned}
& \underset{U\in\mathbb{R}_+^{m\times m}}{\text{minimize}} & &  F(U) \eqdef \big < U U^\top,  \mathbf{1}-2S \big >  \\[-0.75em]
&&&  \qquad\qquad + d\big(\phi_\mathrm{orth}(U) + \phi_\mathrm{dist}(U)\big) \\
& \text{subject to}  & &  U\mathbf{1}_m=\mathbf{1}_m
\end{aligned}.
\end{gather}
\rev{Because $\phi_\mathrm{orth}\ge0$ and $\phi_\mathrm{dist}\ge0$, the parameter $d$ must be non-negative so that any constraint violation incurs a cost.}

\rev{The objective $F(U)$ of problem~\eqref{eq:relaxed} is nonconvex.
Thus, convergence to a first-order stationary point is not enough as critical points may either be local minima or saddle points.
Therefore, higher-order information must be used so that convergence to second-order stationary points can be achieved, implying convergence to local minima~\cite{mokhtari2018escaping}.
We leverage the generic algorithmic framework of~\cite{mokhtari2018escaping} for escaping strict saddle points in constrained optimization by searching for feasible directions in the nullspace of the gradient that have negative curvature with respect to the Hessian $\nabla^2 F$~\cite[Thm.~4]{mokhtari2018escaping}.
Equipped with the ability to converge to second-order stationary points, our main result concerning \eqref{eq:relaxed} is stated in the following theorem, where \emph{solutions} refer to (local) minima.
}

\begin{restatable}{thm}{feasibilityoflocalminima} \label{thm:feasibility-of-local-minima}
\rev{For $d\geq m+1$, solutions $U^\star$ of problem~\eqref{eq:relaxed} are feasible solutions of problem~\eqref{eq:penalty-form}.
In particular, $U^\star$ is cycle consistent, distinct, and binary.}
\end{restatable}
\begin{proof}
See
Appendix~\ref{apdx:analysis}.
\end{proof}

For a comparison of our relaxation and its guarantees with existing approaches having similar formulations, see Table~\ref{tbl:relaxations}.
\rev{The closest formulations to \eqref{eq:relaxed} in Table~\ref{tbl:relaxations} are MatchDGD~\cite{leonardos2018distributed} and MatchRTR~\cite{leonardos2020low}, both of which include a regularizer that penalizes non-binary solutions, similar to the role of $\phi_\mathrm{orth}$.
However, inclusion of this regularizer does not \emph{guarantee} that solutions will be binary.
Thus, a Hungarian-based rounding step is required, incurring additional computation and potentially resulting in a solution that causes an increase in objective.
Additionally, MatchRTR~\cite{leonardos2020low} is not able to guarantee that solutions satisfy distinctness.
}

\textbf{Algorithm.} %
We approach solving \eqref{eq:relaxed} by gradually increasing the scalar parameter $d\geq0$.
The rationale is that the affinity $S$ is indicative of the true solution and so an initially small $d$ allows $U$ to be biased towards a good solution, while a large $d$ ultimately pushes $U$ towards a feasible solution.
According to Theorem~\ref{thm:feasibility-of-local-minima}, the output of MIXER is a cycle-consistent, distinct, and binary second-order (local) minimum of~\eqref{eq:penalty-form}.

MIXER is summarized in Algorithm~\ref{alg:mixer}.
The solution is initialized using the eigenvectors of the modified affinity matrix, $\mathbf{1}-2S$ (Line 3).
\rev{The penalty weight $d$ is first initialized (Line 4) to a value that causes approximately half of the elements of $U$ that violate distinctness or orthogonality to diminish towards zero in the first step (see
Appendix~\ref{apdx:update-rule});
note that $[\cdot]$ refers to an element-wise operation.}
\rev{For each value of $d$, we solve \eqref{eq:relaxed} using projected gradient descent (PGD) with the projection $\Pi_\Delta$ onto the convex constraint set.
The projection operation can be efficiently implemented as a non-iterative algorithm~\cite{wang2013projection}.
PGD over convex constraints is guaranteed to converge to first-order stationary points~\cite[Prop.~7]{mokhtari2018escaping} and the \texttt{DetectAndEscapeSaddle} framework of \cite[Alg.~1]{mokhtari2018escaping} ensures convergence to second-order stationary points, i.e., feasible local minima (cf. Theorem~\ref{thm:feasibility-of-local-minima}).
Note that in practice, we have found that adding a small ($10^{-1}$) perturbation to $P_o$, $P_d$ is efficient to implement and avoids non-binary saddles altogether.
This is because non-binary saddles have strictly negative curvature (cf.
Lemma~4
in
Appendix~\ref{apdx:analysis}),
so moving in random directions is likely to cause objective minimization.
Additionally, we frequently observed that Algorithm~\ref{alg:mixer} converges to feasible solutions of~\eqref{eq:penalty-form} with a much smaller $d$ than required by Theorem~\ref{thm:feasibility-of-local-minima}, and so early termination (Line 11) regularly occurs.}

\begin{algorithm}[t]
\caption{MIXER}
\label{alg:mixer}
\small
\begin{algorithmic}[1]
\State \textbf{Input} affinity matrix $S \in [0,1]^{m\times m}$, set cardinalities $m_i \in \bn^n$ (if distinctness is required) %
\State \textbf{Output} $U \in \{0,1\}^{m\times m}$
\State $U \gets \Pi_\Delta(\mathrm{eigvec}(\mathbf{1} - 2S))$ \algcomment{initialize using eigenvectors}
\State \rev{$d\gets \mathrm{med} \{ - \frac{[\mathbf{1}-2S]}{[UP_o+P_dU]} : [UP_o+P_dU] > 0 , [U] >0 \}$}
\While {$d<m+1$} %
    \While {$U$ not converged}
        \State $\nabla F = 2(\mathbf{1}-2S)U + 2d\big( U P_o + P_d U \big) $
        \State $U \gets \Pi_\Delta(U - \alpha \nabla F)$ \algcomment{$\alpha$ via backtracking line search}
    \EndWhile
    \State \rev{$U \gets \mathrm{\tt DetectAndEscapeSaddle}(U)$}
    \State \rev{$d\gets 2d$}
    \State \algorithmicif\ $\phi_\mathrm{orth}(U)=0$ and $\phi_\mathrm{dist}(U)=0$ \algorithmicthen\ \Return
\EndWhile
\end{algorithmic}
\end{algorithm}

\textbf{Computational complexity.} 
Neglecting constant factors and replacing the \texttt{DetectAndEscapeSaddle} routine with small gradient perturbations as described above, the worst-case complexity of
Algorithm 1 is bounded by $\mathcal{O}(m^3)$ per iteration, corresponding to matrix multiplications. 
We observed that most of the runtime is spent computing the matrix-vector products in $\nabla F$ and $F$ and by the $\mathcal{O}(m^2\log m)$ projection onto the matrix simplex, $\Pi_\Delta$.
A numerical analysis of runtime for different size problems is given in
Appendix~\ref{apdx:timing}).

\section{Experiments}\label{sec:experiments}

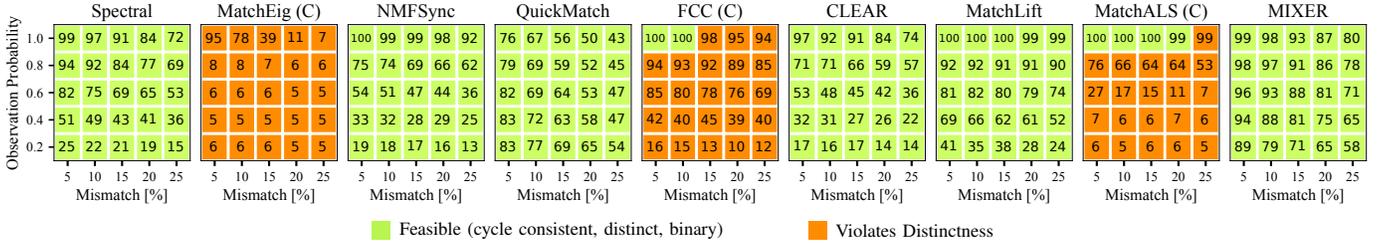
\begin{figure*}[t]
    \centering
    \begin{subfigure}[b]{1\textwidth}
        \includegraphics[trim=0mm 6mm 0mm 0mm, clip, width=1\columnwidth]{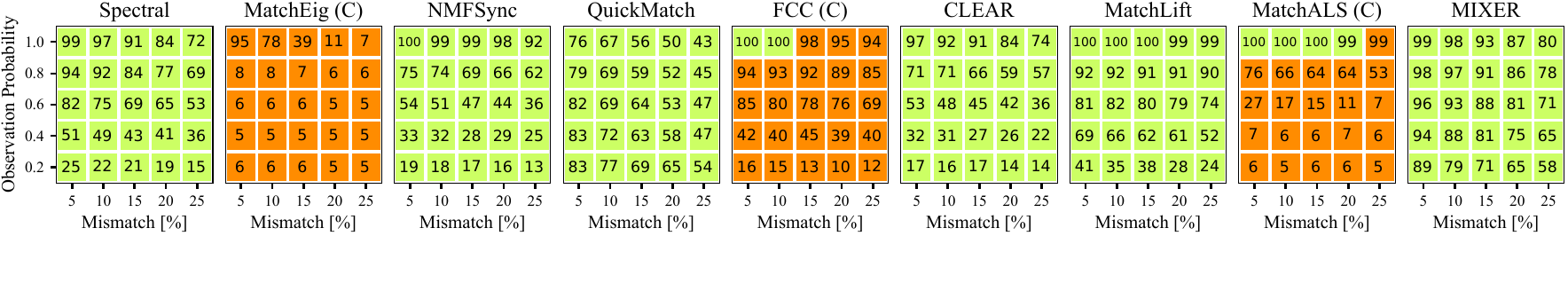}
    \end{subfigure}
    \\
    \vspace{0.025in}
    \begin{subfigure}[b]{1.0\textwidth}
        \centering
        \begin{tikzpicture}
            \node[fill={rgb,255:red,184; green,245; blue,80}, minimum size=2.5mm] (a) at (0,0) {};
            \node[right = -1mm of a] (atxt) {\scriptsize\ Feasible (cycle consistent, distinct, binary)};
            \node[right = of atxt, fill={rgb,255:red,255; green,140; blue,0}, minimum size=2.5mm] (b) {};
            \node[right = -1mm of b] (btxt) {\scriptsize\ Violates Distinctness};
        \end{tikzpicture}
    \end{subfigure}
    \vskip-0.2em
    \caption{
    Multiway early fusion of noisy pairwise affinities generated from $n=10$ simulated views of a universe with $k=30$ distinct objects.
    MIXER accurately fuses pairwise affinities while enforcing cycle consistency and distinctness, performing better than the state-of-the-art over the entire range of noise regimes.
    Especially apparent is the high accuracy of MIXER compared to existing methods in low observation redundancy settings (bottom rows).
    }
    \label{fig:synthetic_results_f1}
\end{figure*}

\begin{figure*}[t]
    \centering
    \begin{subfigure}[b]{1\textwidth}
        \includegraphics[trim=0mm 0mm 0mm 0mm, clip, width=1\columnwidth]{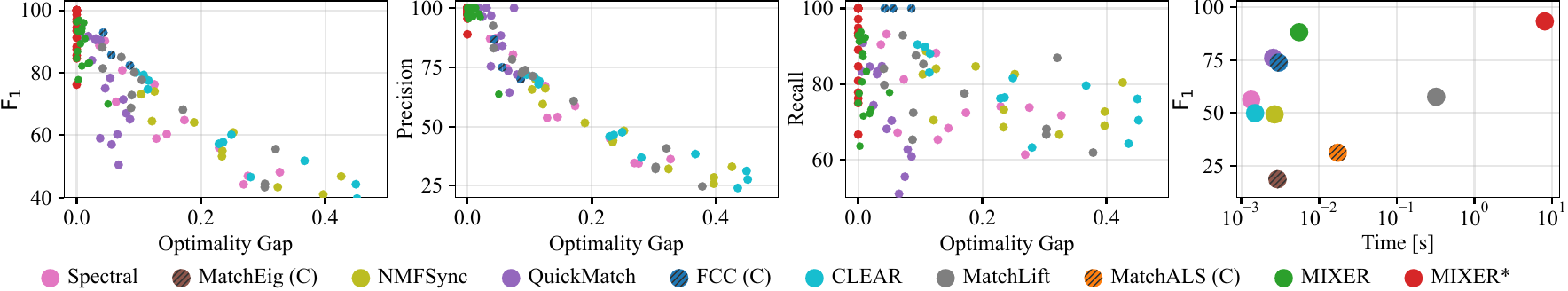}
    \end{subfigure}
    \caption{
    $\mathrm{F}_1$, precision, and recall compared with optimality gaps relative to our MIQP formulation \eqref{eq:miqp}.
    Each point corresponds to a solution of a synthetic problem having some observation probability and mismatch percentage (e.g., see Fig.~\ref{fig:synthetic_results_f1}).
    As the optimality gap of a solution decreases, the accuracy increases, validating our multiway fusion formulation.
    MIXER solutions are tightly grouped in the high-accuracy, near-optimal regime.
    The last plot visualizes each algorithm's average accuracy vs runtime, showing that MIXER is much more scalable than directly solving the MIQP.
    }
    \label{fig:synthetic_results_optgap}
    \vskip-1em
\end{figure*}

\begin{figure}[t]
    \centering
    \includegraphics[trim=0mm 0mm 0mm 0mm, clip, width=1\columnwidth]{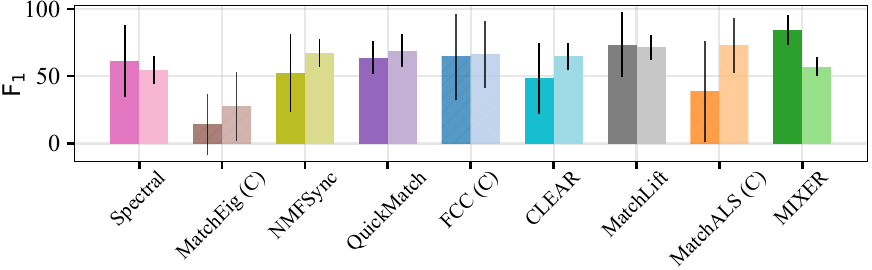}
    \vskip-0.2em
    \caption{
    Average $\mathrm{F}_1$ results using early fusion (left bars) and late fusion (right bars).
    Note that MatchEig (C) and MatchALS (C) violate distinctness constraints and so these infeasible results are indicated with hatch marks.
    }
    \label{fig:synthetic_results_early_late}
\end{figure}

We evaluate MIXER in three experiments.
First, we use synthetic data to validate that our MIQP problem formulation~\eqref{eq:miqp} enables high accuracy.
We find that MIXER achieves near-optimal performance compared to several existing state-of-the-art multiway matching algorithms.
Second, we compare MIXER with other algorithms using standard multiway image feature matching benchmarks.
Finally, we apply multiway fusion to a challenging robotics dataset collected as part of this work, highlighting MIXER's ability to use multiple attributes to improve the overall fusion accuracy.

We report matching accuracy in terms of precision, recall, and the $\mathrm{F}_1$ score.
Precision $p$ is the number of correct associations divided by the total number of associations identified, recall $r$ is the number of correct associations identified divided by the total number of associations in the ground truth, and the $\mathrm{F}_1$ score is defined as $\frac{2pr}{p+r}\in[0,1]$, which captures the balance between precision and recall.

We compare against several state-of-the-art multiway matching algorithms.
Spectral~\cite{Pachauri2013} and MatchEig~\cite{Maset2017} are based on spectral relaxation,
CLEAR~\cite{fathian2020clear} and QuickMatch~\cite{tron2017quickmatch} are based on graph clustering,
FCC~\cite{shi2021fcc} is based on cluster-consistency statistics,
NMFSync~\cite{bernard2019synchronisation} is based on matrix factorization,
MatchALS~\cite{zhou2015multi} is based on low-rank matrix recovery,
and MatchLift~\cite{chen2014near} is based on convex relaxation.
Except for MIXER, QuickMatch and FCC, algorithms require a universe size estimate and we use CLEAR's estimate as the typical spectral approach is typically sensitive to noisy affinity matrices~\cite{fathian2020clear,zhou2015multi,Maset2017}.
For MatchALS, we scale this estimate by $2$ as suggested by the authors.
As a baseline, we perform pairwise data association using Hungarian with a minimum score threshold ($0.35$), as commonly done in the tracking literature~\cite{bewly2016sort}, demonstrating the type of noisy, cycle inconsistent results that arise when naively fusing a set of pairwise associations.
\rev{For algorithms that do not guarantee cycle consistency (baseline, FCC, MatchEig, MatchALS), we post-process their solutions by extracting connected components and completing them into cliques (cf.~\cite[\S VII]{fathian2020clear}).
This enforces cycle consistency, but exposes distinctness violations---thus highlighting the importance of simultaneously satisfying both constraints during solution search.
These completed results are denoted with a ``(C)''.}
For small problems, we globally solve the MIQP \eqref{eq:miqp} to optimality using Gurobi~9.5.2, referred to as \mixeropt. %
All algorithms are implemented in MATLAB and executed on an i7-6700 CPU with 32 GB RAM.

\subsection{Synthetic Dataset}\label{sec:synthexpts}
We use Monte Carlo analysis with synthetic data to compare MIXER with several state-of-the-art algorithms across different noise regimes.
We aim to 1) validate our formulation \eqref{eq:miqp} for high-accuracy multiway fusion, 2) show the ability of MIXER to obtain near-optimal and thus high-accuracy solutions, 3) show that early fusion with MIXER achieves higher accuracy than late fusion with existing methods, and 4) show that MIXER provides a computationally scalable approach to approximately solve \eqref{eq:miqp}.
Noisy pairwise associations are synthetically generated by considering $n=10$ partial views of $k=30$ objects, e.g., $10$ images each containing at most $30$ objects.
Twenty-five different noise regimes are considered, wherein the percent mismatch (i.e., binary noise added to ground truth associations) and the probability of observation (i.e., selection of random subsets of the universe) are varied.
A noisy association $a^{ij}_{kl}\in\{0,1\}$ of objects $k$ and $l$ from views $i$ and $j$ is then made into an affinity $s^{ij}_{kl}\in[0,1]$ by adding uncertainty according to $s(a; \theta) = (1 - \theta)a + 0.5\theta$, where $\theta$ is drawn from a standard uniform distribution.
Each algorithm processes these noisy pairwise affinities, allowing us to study the accuracy of each method as an early fusion approach.

Fig.~\ref{fig:synthetic_results_f1} shows average results over 10 Monte Carlo trials.
While all algorithms generally perform well in low noise regimes with perfect observability (top left squares of Fig.~\ref{fig:synthetic_results_f1}), the performance of most algorithms quickly deteriorates as the probability of observing objects decreases.
MIXER is the exception, performing better than other approaches on average, over the entire range of mismatch and partial observability.
The low observation redundancy regimes (bottom rows in Fig.~\ref{fig:synthetic_results_f1}) are especially important because these settings occur when limited field-of-view sensors only see portions of the universe; for example, in multirobot SLAM~\cite{aragues2011consistent,tian2022kimera}.

To assess how well our formulation \eqref{eq:miqp} addresses the multiway fusion problem, we compute the optimality gap of each algorithm's solutions compared to \mixeropt{}. %
Fig.~\ref{fig:synthetic_results_optgap} plots the accuracy of each solution against the optimality gap, exposing the correlation between low optimality gap and high accuracy ($F_1$ and precision).
MIXER solutions are tightly grouped in this near-optimal, high-accuracy regime, showing that our algorithm, although a local, first-order method, is frequently able to achieve good results.
Interestingly, we observe in the recall results of Fig.~\ref{fig:synthetic_results_optgap} that our formulation leads MIXER to be conservative---it would rather \emph{not} fuse objects if precision would be sacrificed.
This property allows further improvements to be made as more data is collected.

Finally, Fig.~\ref{fig:synthetic_results_early_late} highlights that early fusion with MIXER outperforms both early and late fusion of uncertain affinities compared to existing state-of-the-art algorithms.
In late fusion, pairwise affinities are first processed into hard matches before multiway matching is performed, and in this setting, MIXER does not perform as well.
The principal reason is because MIXER is a multiway MWM formulation as discussed in Remark~\ref{rem:multiway-mwm}---when pairwise associations are first made, a large amount of information is discarded, precluding MIXER from using it in combination with multiway constraints.

\subsection{Benchmark Datasets}
We use the
CMU Hotel\footnote{ \href{http://pages.cs.wisc.edu/~pachauri/perm-sync/}{http://pages.cs.wisc.edu/\textasciitilde pachauri/perm-sync/}}
and
Affine Covariant Features\footnote{\href{https://www.robots.ox.ac.uk/~vgg/data/affine}{https://www.robots.ox.ac.uk/\textasciitilde vgg/data/affine}}
benchmark datasets to demonstrate multiway fusion with MIXER on real data and report results in Table~\ref{tbl:benchmark_results}.
These datasets consist of collections of images from different perspectives and the goal is to extract features from each image, perform pairwise feature matching, and then perform multiway fusion of features.
Instead of standard pairwise SIFT matching, which would cause late fusion since it makes hard decisions about pairwise association and thus limits the ability of multiway fusion to make fully-informed decisions, we instead construct pairwise affinity matrices using k-nearest neighbors.
Pairwise affinities are created by finding the $10$ closest SIFT matches ($\ell_1$ distance) for each keypoint.
The affinity score of the closest neighbor is set to $1$, the other neighbors set to $0.5$, and the rest to $0$.
This makes use of all three states of association and indicates that for the semi-close neighbors, there is some uncertainty as to whether or not they should be matched, while for far away neighbors they are very likely to be incorrect matches.

\begin{table}[t]
\scriptsize
\centering
\caption{
Benchmark results on CMU Hotel and Affine Covariant Features datasets.
$\mathrm{F}_1$ scores are reported as percentages.
Results that violate cycle consistency are indicated in {\color{vred}red} and are completed to asses their true fusion accuracy.
When fusing cycle inconsistent solutions, distinctness is often violated and these results are indicated in {\color{vorange}orange}.
In every case, MatchLift is at least 1 order of magnitude slower than MIXER (see Fig.~\ref{fig:synthetic_results_optgap}).
}
\setlength{\tabcolsep}{1pt}
\ra{1.1}
\sisetup{detect-weight=true,detect-inline-weight=math}
\begin{tabular}{@{}l S S S S S S S S S@{}}
\toprule
Algorithm & {Hotel} & {Wall} & {UBC} & {Bikes} & {Leuven} & {Trees} & {Graffiti} & {Bark} & {Boat} \\
\toprule
Baseline
    & \color{vred}89.2 & \color{vred}49.0 & \color{vred}71.1 & \color{vred}52.4 & \color{vred}68.9 & \color{vred}44.0 & \color{vred}40.9 & \color{vred}39.4 & \color{vred}47.6 \\
Baseline (C)
    & \color{vorange}16.8 & \color{vorange}2.2 & \color{vorange}5.9 & \color{vorange}1.4 & \color{vorange}4.9 & \color{vorange}1.5 & \color{vorange}0.3 & \color{vorange}10.7 & \color{vorange}0.9 \\
Spectral
    & 65.3 & 43.8 & 51.5 & 48.5 & 58.5 & 46.2 & 34.1 & 31.4 & 26.8 \\
MatchEig
    & \color{vred}94.7 & \color{vred}78.4 & \color{vred}84.4 & \color{vred}75.7 & \color{vred}78.9 & \color{vred}50.4 & \color{vred}48.0 & \color{vred}37.5 & \color{vred}61.2 \\
MatchEig (C)
    & \color{vorange}27.2 & \color{vorange}16.1& \color{vorange}10.8 & \color{vorange}1.9 & \color{vorange}4.1 & \color{vorange}1.6 & \color{vorange}0.3 & \color{vorange}4.0 & \color{vorange}1.1 \\
NMFSync
    & 89.7 & 39.4 & 63.2 & 58.4 & 59.5 & 33.5 & 29.3 & 18.9 & 44.1 \\
QuickMatch
    & 82.4 & 46.4 & 65.7 & 50.9 & 62.7 & 42.3 & 39.5 & 28.2 & 43.4 \\
FCC
    & \color{vred}92.5 & \color{vred}55.7 & \color{vred}77.4 & \color{vred}59.0 & \color{vred}68.8 & \color{vred}51.7 & \color{vred}40.4 & \color{vred}42.1 & \color{vred}32.3 \\
FCC (C)
    & \color{vorange}92.8 & \color{vorange}46.3 & \color{vorange}70.8 & \color{vorange}52.8 & \color{vorange}58.5 & \color{vorange}41.0 & \color{vorange}32.4 & \color{vorange}30.2 & \color{vorange}29.9 \\
CLEAR
    & 76.9 & 40.9 & 47.8 & 40.3 & 53.1 & 26.1 & 28.1 & 25.8 & 38.9 \\
MatchALS
    & \color{vred}95.4 & \color{vred}71.4 & \color{vred}77.1 & \color{vred}71.1 & \color{vred}73.9 & \color{vred}50.8 & \color{vred}46.7 & \color{vred}38.4 & \color{vred}58.7 \\
MatchALS (C)
    & \color{vorange}89.3 & \color{vorange}43.1 & \color{vorange}26.4 & \color{vorange}39.2 & \color{vorange}39.2 & \color{vorange}21.9 & \color{vorange}11.1 & \color{vorange}26.9 & \color{vorange}35.0 \\
MatchLift
    & 95.9 & 55.9 & \bfseries 79.2 & 63.8 & 71.9 & 48.2 & \bfseries 45.4 & 35.0 & 54.4 \\
\midrule
MIXER
    & \bfseries 97.0 & \bfseries 56.4 & 77.7 & \bfseries 67.5 & \bfseries 74.2 & \bfseries 51.0 & 43.3 & \bfseries 40.0 & \bfseries 59.2 \\
\bottomrule
\end{tabular}
\label{tbl:benchmark_results}
\end{table}

\subsection{Car Fusion Dataset}\label{sec:carexpt}
\begin{figure}[t]
    \centering
    \includegraphics[width=1\columnwidth]{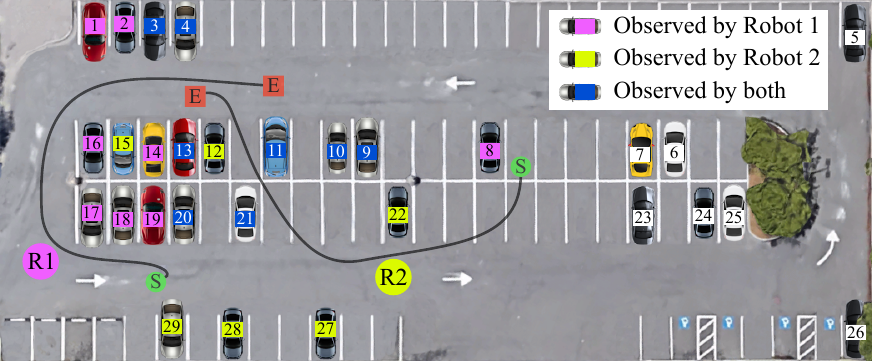}
    \caption{
    Illustration of the parking lot dataset, with the paths of two robots (R1 and R2), from start (S) to end (E).
    The robots observed $k=22$ cars, with $8$ cars being covisible. 
    Colors indicate which robot saw which car.
    }
  \label{fig:zpark}
\end{figure}

In this section, we evaluate MIXER's ability to fuse observation of objects seen from multiple views.
Multiway object fusion is a task that appears in settings like multirobot SLAM~\cite{aragues2011consistent,tian2022kimera} and multimodal object tracking~\cite{chiu2021probabilistic}.
We collect data by teleoperating a Clearpath Jackal equipped with an RGB camera around a parking lot as shown in Fig.~\ref{fig:zpark}.
We select $n=184$ images from two separate traversals to create large view changes and thus increase the difficulty.
In total, there are $k=22$ distinct cars visible from $339$ car detections, resulting in only $8\%$ of the universe being observed in each image.
Using these RGB images, the objective is to fuse cars seen from different views using noisy, partial, and cropped detections (i.e., not every car is seen in each frame, and some cars extend out of frame).
Cars are detected using YOLOv3~\cite{redmon2018yolov3} and we extract three attributes from each detection: bounding box, car color, and visual appearance.
These three attributes have complementary strengths. %
For example, bounding boxes are good for scoring consecutive detections, but cannot be used over large time offsets; car color is a sensitive quantity to extract, but is good for associating detections over large time offsets; visual appearance based on the number of matching SIFT descriptors is typically robust to small/medium view changes, but breaks down with large view changes.
Thus, using these attributes highlights the three-states-of-association feature of the formulation of problem~\eqref{eq:miqp}, which allows MIXER to achieve high accuracy.
Ground truth associations are generated by manually annotating the detections.

Each attribute creates its own set of pairwise affinities, which will be combined to create a multiway affinity matrix.
Bounding box affinity is scored using intersection-over-union (IoU) for consecutive detections, while non-consecutive detections take a value of $0.5$ as wide-baseline IoU scoring is inconclusive.
Color affinity scores take either $0$, $0.5$, or $1$, with $0.5$ being used if either car's color could not be clearly ascertained.
Visual similarity is scored based on the number of SIFT matches between two cars.
Common attribute affinity combination/fusion methods include (weighted) averaging, non-maximum suppression, probabilistic ensembling~\cite{chen2022multimodal}, multi-layer perceptrons~\cite{chiu2021probabilistic}, graph neural networks~\cite{weng2020gnn3dmot}, or other learned fusion processes~\cite{zhang2019robust}.
For simplicity, we adopt the weighted averaging approach, using a weights of $1$, $0.5$, and $1$ for bounding box overlap, color similarity, and appearance similarity, respectively.
Color receives less weight due to its lack of robustness.

\begin{table}[t]
\scriptsize
\centering
\caption{
Multiway car fusion results.
Objective values are listed except for solutions that violate cycle consistency ({\color{vred}red}) or distinctness ({\color{vorange}orange}).
}
\setlength{\tabcolsep}{6pt}
\ra{1.1}
\sisetup{detect-weight=true,detect-inline-weight=math}
\begin{tabular}{@{}l S S S S H S[table-format=1,table-alignment=center]@{}}
\toprule
Algorithm & {Precision} & {Recall} & {$\mathrm{F}_1$} & {Obj. $(\downarrow)$} & $\hat{k}$ & {Time [ms]} \\
\toprule
Baseline
    & \color{vred}44.9 & \color{vred}14.8 & \color{vred}22.3 & {--} & 44 & 1648 \\
Baseline (C)
    & \color{vorange}8.2 & \color{vorange}84.5 & \color{vorange}14.9 & {--} & 44 & 1650 \\
Spectral
    & 22.8 & 58.0 & 32.7 & 146.9 & 11 & \bfseries 12 \\
MatchEig
    & \color{vred}13.0 & \color{vred}72.8 & \color{vred}22.1 & {--} & 2 & 1079 \\
MatchEig (C)
    & \color{vorange}6.3 & \color{vorange}99.7 & \color{vorange}11.8 & {--} & 2 & 1083 \\
NMFSync
    & 15.7 & 58.6 & 24.7 & 151.5 & 5 & 35 \\
QuickMatch
    & 13.8 & 31.2 & 19.2 & 148.9 & 68 & 37 \\
FCC
    & \color{vred}47.4 & \color{vred}70.8 & \color{vred}56.8 & {--} & 7 & 3269 \\
FCC (C)
    & \color{vorange}6.3 & \color{vorange}98.1 & \color{vorange}11.9 & {--} & 7 & 3272 \\
CLEAR
    & 19.0 & 66.4 & 29.6 & 150.6 & 5 & 63 \\
MatchALS
    & \color{vred}7.5 & \color{vred}94.9 & \color{vred}13.9 & {--} & 1 & 542 \\
MatchALS (C)
    & \color{vorange}6.3 & \color{vorange}99.9 & \color{vorange}11.8 & {--} & 1 & 545 \\
MatchLift
    & 12.4 & 48.8 & 20.1 & 149.7 & 44 & 15463 \\
MatchALS$_{\alpha=0.5}$
    & \color{vred}41.9 & \color{vred}76.7 & \color{vred}54.2 & {--} & 3 & 614 \\
MatchALS$_{\alpha=0.5}$ (C)
    & \color{vorange}8.9 & \color{vorange}87.6 & \color{vorange}16.1 & {--} & 3 & 617 \\
MatchLift$_{\lambda=0.5}$
    & 34.6 & 54.7 & 42.4 & 141.1 & 85 & 15295 \\
\midrule
MIXER
    & \bfseries 79.4 & \bfseries 70.6 & \bfseries 74.8 & \bfseries 135.7 & 57 & 312 \\
\bottomrule
\end{tabular}
\label{tbl:carfusion_results}
\end{table}

\begin{figure}[t!]
    \centering
    \includegraphics[clip, width=1\columnwidth]{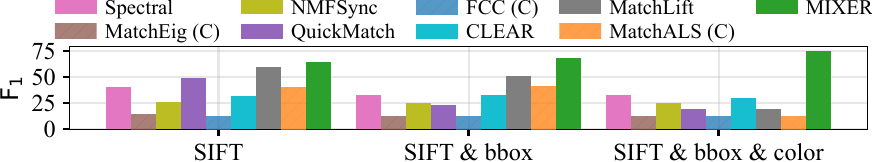}
    \caption{
    Multiway car fusion results, incrementally adding additional attribute affinities.
    Mixing together attributes allows MIXER to increase in $\mathrm{F}_1$ score by more than $10\%$.
    In contrast, the combination of additional uncertain information causes other methods to decrease in $\mathrm{F}_1$ accuracy.
    }
    \label{fig:attribute_ablation}
    \vskip-0.045in
\end{figure}

We also report results for MatchALS$_{\alpha=0.5}$ and MatchLift$_{\lambda=0.5}$ with parameters set so that the input affinity matrix is transformed to $\mathbf{1}-2S$ like in MIXER (see Table~\ref{tbl:relaxations}).
In MatchALS and MatchLift, these parameters---and therefore the coefficient scaling on $S$---are heuristically introduced to encourage sparsity, with default values suggested by their authors of $\alpha=0.1$ and $\lambda=\sqrt{\binom{n}{2}/(2n)}\approx0.3$, respectively.
In contrast, the expression $\mathbf{1}-2S$ arises naturally in our formulation due to the Frobenius objective~\eqref{eq:miqp} and our insight is that it allows for three states of association in multiway MWM (see Remark~\ref{rem:multiway-mwm}).
Setting $\alpha=\lambda=0.5$ allows for a direct comparison of the relaxations and algorithms listed in Table~\ref{tbl:relaxations} and, as expected due to our insights, they provide an increase in accuracy. %
However, as reported in Table~\ref{tbl:carfusion_results}, MIXER substantially outperforms existing algorithms.
It is followed by MatchLift$_{\lambda=0.5}$, which lags by $32\%$ in $\mathrm{F}_1$ accuracy and is $49$x slower.
Although MatchALS$_{\alpha=0.5}$ and MatchLift$_{\lambda=0.5}$ also have three states of association, their relaxations require rounding to binary matrices and search for solutions in association space $A\in\mathbb{R}_+^{m\times m}$, leading to sensitivity of the universe size estimate.

\rev{MIXER achieves the best objective and $\mathrm{F}_1$ accuracy in Table~\ref{tbl:carfusion_results}.
This is consistent with Fig.~\ref{fig:synthetic_results_optgap}, which shows that MIXER attains near-optimal, high-accuracy solutions, even in low-observation-probability settings.
The relationship between near-optimal and high accuracy solutions is due to our MIQP formulation~\eqref{eq:miqp}, which allows for three states of association and therefore a way to combine uncertain affinities from multiple attributes, as shown in Fig.~\ref{fig:attribute_ablation}.}

\section{Conclusion}\label{sec:conclusion}

We presented the MIXER algorithm for multiattribute, multiway fusion of uncertain pairwise affinities.
Our MIQP formulation leverages direct access to affinities and allows three modes of association (non-match, undecided, and match) over the range $0$ to $1$.
This feature led to the insight that our formulation is a multiway extension of the maximum-weight matching problem (see Remark~\ref{rem:multiway-mwm}).
Because of scalability issues of solving MIQPs, we proposed a novel continuous relaxation in a projected gradient descent scheme to converge to feasible, binary solutions of the original problem.
The guarantee of our algorithm to converge to binary points sets it apart from related work, which relies on a final binarization step that can lead solutions to be infeasible.
Finally, our experimental evaluations in three datasets showed that MIXER frequently achieves higher accuracy than the state of the art, especially in noisy regimes with low observation redundancy.
These properties and results establish MIXER as an effective and scalable multiway fusion algorithm in the presence of uncertain affinities.

\bibliographystyle{IEEEtran}
\bibliography{refs}

\appendices

\section{Equivalent Penalty Form} \label{apdx:equiv-penalty-form}

We show step-by-step derivations leading to an equivalent penalty form of problem~\eqref{eq:miqp}, repeated below for convenience
\begin{gather*} %
\begin{aligned}
& \underset{U \in \{0, 1\}^{m\times m}}{\text{minimize}} & &  \big\| UU^\top -  S \big\|_{F}^2 & {\color{gray} \text{(cycle consistency)}}  
\\
& ~~\text{subject to}  & &  U \, \mathbf{1}_m  = \mathbf{1}_m  & {\color{gray} \text{(one-to-one constraint)}} 
\\
&&&  U_i^\top \mathbf{1}_{m_i} \leq \mathbf{1}_m &  {\color{gray} \text{(distinctness constraint)}}
\end{aligned}
\end{gather*}

\noindent \textbf{Equivalent problem.}
From the definition of Frobenius norm, the objective of \eqref{eq:miqp} can be expanded and simplified as
\begin{align} \label{eq:misc1}
\|UU^\top - S\|^2_F
&=
\big< UU^\top - S,~ UU^\top - S \big> \nonumber \\
&=
{\|UU^\top \|^2_F} - {2 \big< UU^\top ,~ S \big>} + {\| S \|^2_F},
\end{align}
where $ \big< A,B \big> \eqdef \tr(A^\top B) = \sum_{ij}{A_{ij} B_{ij}}$ is the Frobenius inner product of matrices $A$ and $B$. 
Further, it holds that 
\begin{equation} \label{eq:misc2}
\|UU^\top \|^2_F
= \sum_{i,j}{(UU^\top)_{ij}^2}
= \sum_{i,j}{(UU^\top)_{ij}}
= \big< UU^\top, \mathbf{1}_{m\times m} \big >,
\end{equation}
where the second equality follows by noting that entries of $U$ are binary and therefore equal to their square.
Combining \eqref{eq:misc1} and \eqref{eq:misc2}, the objective can be written as 
\begin{equation} \label{eq:misc3}
\| UU^\top -  S \|^2_F
=
\big< UU^\top,~ \mathbf{1}_{m\times m} - 2S \big> + {\| S \|^2_F}.
\end{equation}
Since $S$ is given and is a constant of the optimization problem, the term $\|S \|^2_F$ does not affect the solution and can be omitted from the objective.
Thus, defining ${\bar{S} \eqdef \mathbf{1}_{m\times m} - 2S}$, problem~\eqref{eq:miqp} is equivalent to the following problem
\begin{gather} \label{eq:innerProd}
\begin{aligned}
& \underset{U \in \{0, 1\}^{m\times m}}{\text{minimize}} & &  \big< UU^\top,  \bar{S} \big> & {\color{gray} \text{(cycle consistency)}} \\
& ~\text{subject to}
 &&  U\,\mathbf{1}_m = \mathbf{1}_m  &  {\color{gray} \text{(one-to-one constraint)}} \\
&&&  U_i^\top \mathbf{1}_{m_i} \leq \mathbf{1}_m & {\color{gray} \text{(distinctness constraint)}}
\end{aligned}
\end{gather}

\noindent \textbf{Penalty functions.}
Toward deriving the proposed relaxation, we introduce the standard simplex and two penalty functions, allowing the reformulation of the constraints of \eqref{eq:innerProd} into an equivalent problem.
With slight abuse of notation, the \emph{standard simplex} defined for each row of a matrix is
\begin{align}\label{eq:matrix-simplex}
\Delta^{m\times m} \eqdef \{U\in\mathbb{R}_+^{m\times m} : U\mathbf{1}_m=\mathbf{1}_m\}.
\end{align}
\rev{Observe that $\{0,1\}^{m\times m}\subset\Delta^{m\times m}$ and that $\Delta^{m\times m}$ captures the one-to-one constraint.}
Using the standard simplex, we first show that a binary $U\in\Delta^{m\times m}$ must have orthogonal columns.
This orthogonality property is \textit{implicit} in the original formulation and we explictly include it in the relaxation.

\begin{restatable}{lemma}{binaryisorthogonal} \label{lem:binary-is-orthogonal}
A matrix $U\in\Delta^{m\times m}$ is binary if and only if the columns of $U$ are orthogonal.
\end{restatable}
\begin{proof} 
Suppose, by contradiction, two columns $v$ and $w$ of $U$ give $\big<v, w\big> \neq 0$.
Then, since $U$ is binary, there exists at least one $k \in \{1,\dots,m\}$ such that $v_k = 1$ and $w_k = 1$. 
This implies that there are at least two $1$ entries in the $k$'th row of $U$.
Consequently, the $k$-th row of vector $U\mathbf{1}_m$ is strictly greater than $1$, which violates the constraint $U\mathbf{1}_m = \mathbf{1}_m$.

Conversely, assume $U\in\Delta^{m\times m}$ has orthogonal columns.
Without loss of generality, we show that the first row of $U$ is binary and, by applying a similar argument to other rows, we conclude that $U$ is binary.
Denote the first row of $U$ by $v_1$.
Because $U\in\Delta^{m\times m}$, there exists $i\in\{1,\dots,m\}$ such that $(v_1)_i>0$.
Let $u_i$ denote the $i$-th column of $U$.
Orthogonality of columns implies that for all $j\neq i$,
\begin{gather}\label{eq:binary-is-orthogonal-sum}
\begin{aligned}
0
&= u_i^\top u_j = \sum^m_{k=1} (u_i)_k (u_j)_k \\
&= (u_i)_1 (u_j)_1 + \sum^m_{k=2} (u_i)_k (u_j)_k \\
&= (v_1)_i (v_1)_j + \sum^m_{k=2} (u_i)_k (u_j)_k.
\end{aligned}
\end{gather}
Because $U$ is non-negative, for \eqref{eq:binary-is-orthogonal-sum} to hold we must have $(v_1)_i(v_1)_j=0$ for all $j\neq i$.
Since $(v_1)_i>0$, it follows that $(v_1)_j=0$ for all $j\neq i$.
Further, since the sum of the entries of $v_1$ is equal to $1$, we have that $(v_1)_i=1$ so that the first row of $U$ is binary.
\end{proof}

Because we seek binary $U\in\Delta^{m\times m}$, we introduce a penalty function corresponding to the column-wise orthogonality of $U$, defined as
\begin{equation} \label{eq:penalty-orth}
\phi_\mathrm{orth}(U) \eqdef \big<  U^\top U,~ P_o\big>,
\end{equation}
where $P_o \eqdef \mathbf{1}_{m\times m} - I_{m\times m}$.
Note that by definition, $\big<  U^\top U,~ P_o\big> \eqdef \sum_{i,j}{ (U^\top U)_{ij} \, (P_o)_{ij}} = \sum_{i\neq j}{ (U^\top U)_{ij} }$, which is the sum of non-diagonal entries.
Therefore, $\phi_\mathrm{orth}(U) = 0$ if and only if the columns of $U$ are orthogonal.

The second penalty function corresponds to the distinctness constraint and is defined as 
\begin{equation} \label{eq:penalty-dist}
\phi_\mathrm{dist}(U) \eqdef \big< UU^\top,~ P_d \big>,
\end{equation}
where $P_d \eqdef \mathrm{blockdiag}(P_{d1}, \dots, P_{dn})$ and each $m_i\times m_i$ matrix $P_{di} \eqdef 2\,(\mathbf{1}_{m_i \times m_i} - I_{m_i \times m_i})$ ensures that the $m_i$ observations of view $i$ are distinct.

\begin{restatable}{lemma}{distinctness} \label{lem:distinctness}
Given $U \in \{0,1\}^{m\times m}$ and $\phi_\mathrm{dist}(U)$ as defined, $\phi_\mathrm{dist}(U) = 0$ if and only if $U_i^\top \mathbf{1}_{m_i} \leq \mathbf{1}_m$. 
\end{restatable}
\begin{proof} 
Expanding $\phi_\text{dist}(U)$ based on the $n$ matrix blocks in $U$ and $P_d$ gives
\begin{gather}
\begin{aligned}
\big< UU^\top,~ P_d \big>
&= {\sum_{i=1}^{n}{\big< U_iU_i^\top,~ P_{di} \big>}} \\
&= {\sum_{i=1}^{n}{\big< U_iU_i^\top,~ \mathbf{1}_{m_i \times m_i} - I_{m_i \times m_i} \big>}} \\
&= {\sum_{i=1}^{n}{\sum_{j\neq r}{(U_i U_i^\top)_{jr}}}}.
\end{aligned}
\end{gather}
Since $U_i$ is binary, the latter summation is zero if and only if all matrices $U_i U_i^\top$ are diagonal.
Suppose $\big< U U^\top,~ P_d \big> = 0$ and, by contradiction, there exists $U_i$ for which $U_i U_i^\top$ is non-diagonal.
This implies $U_i$ has at least two non-orthogonal rows.
From a similar proof-by-contradiction argument used in the proof of Lemma~\ref{lem:binary-is-orthogonal} (based on rows instead of columns), non-orthogonality implies that there exists $k$ such that the $k$-th elements of these two non-orthogonal rows are $1$.
Therefore, the $k$-th element of $U_i^\top \mathbf{1}_{m_i}$ is strictly greater than $1$, a contradiction.
Now suppose $U_i^\top \mathbf{1}_{m_i} \leq \mathbf{1}_m$. 
Since $U_i$ is binary, this implies that if the $k$-th element of a row of $U_i$ is $1$, then the $k$-th element of all other rows must be $0$. Consequently, rows of $U_i$ are orthogonal, which implies ${\sum_{j\neq r}{(U_i U_i^\top)_{jr}}} = 0$ and therefore $\big< UU^\top,~ P_d \big> = 0$.
\end{proof}

Using the standard simplex \eqref{eq:matrix-simplex} and the penalty functions \eqref{eq:penalty-orth}, \eqref{eq:penalty-dist}, problem \eqref{eq:innerProd} can be equivalently expressed as problem~\eqref{eq:penalty-form}, which is repeated below for convenience
\begin{gather*} %
\begin{aligned}
& \underset{U}{\text{minimize}} & &  \big < U U^\top,  \bar{S} \big > & {\color{gray} \text{(cycle consistency)}} \\
& \text{subject to}
 &&  U\in\Delta^{m\times m}  &  {\color{gray} \text{(one-to-one constraint)}} \\
&&&  U\in\{0,1\}^{m\times m}  &  {\color{gray} \text{(binary constraint)}} \\
&&&  \phi_\mathrm{orth}(U) = 0  &  {\color{gray} \text{(orthogonality constraint)}} \\
&&&  \phi_\mathrm{dist}(U) = 0 & {\color{gray} \text{(distinctness constraint)}}
\end{aligned}
\end{gather*}

\section{Theoretical Analysis}\label{apdx:analysis}

We present theoretical insights behind the relaxed problem~\eqref{eq:relaxed} which lead to the MIXER algorithm.
Consider the relaxation in standard form \cite{nocedal1999numerical} so that the constraints are explicit and with $\bar{S}\eqdef\mathbf{1}-2S$, restated here for convenience
\begin{align} \label{eq:relaxed-standard-form}
& \underset{U\in\mathbb{R}_+^{m\times m}}{\text{minimize}} & &  F(U) \eqdef \big < U U^\top,  \bar{S} \big > + d\big(\phi_\mathrm{orth}(U) + \phi_\mathrm{dist}(U)\big) \notag \\
& \text{subject to}
 &&  U \geq \mathbf{0}_{m\times m} \\
&&&  U \mathbf{1}_{m} - \mathbf{1}_{m} = \mathbf{0}_{m} \notag
\end{align}
\rev{The scalar $d\geq0$ controls the strength of the penalty functions.
Intuitively, increasing $d$ pushes solutions of \eqref{eq:relaxed-standard-form} towards binary, distinct solutions.
Because $\phi_\mathrm{orth}(U)\ge0$ and $\phi_\mathrm{dist}(U)\ge0$, $d$ is restricted to be non-negative so that constraint violation penalizes the objective.
In this appendix, we will show that using Algorithm~\ref{alg:mixer}, once $d$ is larger than a finite value $\phi_\mathrm{orth}(U)=\phi_\mathrm{dist}(U)=0$ so that local minima of \eqref{eq:relaxed-standard-form} are feasible solutions of problem~\eqref{eq:penalty-form}.}

The Lagrangian of \eqref{eq:relaxed-standard-form} is
\begin{equation} \label{eq:relaxed-lagrangian}
\mathcal{L}(U; Y,\lambda) \eqdef F - \big< Y,~ U\big> - \big<\lambda,~ U \mathbf{1}_m - \mathbf{1}_m \big>,
\end{equation}
where $\lambda\in\mathbb{R}^m$ and $Y\in\mathbb{R}^{m\times m}$ are the Lagrange multipliers for the equality and inequality constraints, respectively.
From the first-order optimality conditions, stationary points ${U^\star\in\Delta^{m\times m}}$ of \eqref{eq:relaxed-standard-form} must satisfy the KKT conditions
\begin{subequations} \label{eq:kkt}
\begin{align}
\nabla_U\mathcal{L} = \nabla F(U^\star) - Y^\star - \lambda^\star\mathbf{1}_m^\top &= 0, \label{eq:kkt-stationarity} \\
U^\star \mathbf{1}_{m} - \mathbf{1}_{m} &= \mathbf{0}_{m}, \\
U^\star_{ij} &\geq 0, \;\forall_{ij}, \\
Y^\star_{ij} &\geq 0, \;\forall_{ij}, \label{eq:kkt-dual-feasibility} \\
Y^\star_{ij}\,U^\star_{ij} &=0, \;\forall_{ij}, \label{eq:kkt-complementarity}
\end{align}
\end{subequations}
where $\nabla F(U)=2\bar{S}U + 2d\big( U P_o + P_d U \big)$.

\rev{We begin by analyzing gradient entries corresponding to non-zero entries of a given stationary point.}

\begin{lemma} \label{lem:u-zero-or-gradf-equal}
At a stationary point $U^\star\in\Delta^{m\times m}$ of \eqref{eq:relaxed-standard-form}, entries $U^\star_{ij}\neq0$ of the $i$-th row have equal corresponding gradient entries $\nabla F_{ij}$.
In particular, ${\nabla F_{ij} = \sum_{k=1}^m (\nabla F \odot U^\star)_{ik}}$.
\end{lemma}
\begin{proof}
Considering the stationarity condition \eqref{eq:kkt-stationarity}, we can remove the dependence on $Y^\star$ by element-wise right multiplication of $U^\star$.
Because of complementarity \eqref{eq:kkt-complementarity}, we have
\begin{equation}\label{eq:kkt-stationarity-noY}
\nabla F \odot U^\star - \lambda^\star\mathbf{1}_m^\top\odot U^\star = 0,
\end{equation}
where $\odot$ denotes the Haddamard product.
Observing that $(\lambda^\star\mathbf{1}_m^\top\odot U^\star)\mathbf{1}_m=\lambda^\star$ because $U^\star\mathbf{1}_m=\mathbf{1}_m$, multiplying on the right by~$\mathbf{1}_m$ and isolating $\lambda^\star$ yields
\begin{equation}\label{eq:isolated-lambda}
\lambda^\star = (\nabla F \odot U^\star)\mathbf{1}_m
\end{equation}
Our aim is to eliminate explicit dependence of $\lambda^\star$ from \eqref{eq:kkt-stationarity-noY}.
Substituting \eqref{eq:isolated-lambda} into \eqref{eq:kkt-stationarity-noY} yields
\begin{align}
\nabla F \odot U^\star - (\nabla F \odot U^\star)\mathbf{1}_m\mathbf{1}_m^\top\odot U^\star &= 0 \notag \\
(\nabla F - (\nabla F \odot U^\star)\mathbf{1}_{m\times m})\odot U^\star &= 0, \notag
\end{align}
which implies one of the following two cases
\begin{equation*}
\begin{cases}
U^\star_{ij} = 0 \\
U^\star_{ij} \neq 0,\
(\nabla F - (\nabla F \odot U^\star)\mathbf{1}_{m\times m})_{ij} = 0
\end{cases}
\end{equation*}
Through simplification, when $U^\star_{ij}\neq0$ we have
\begin{align}\label{eq:gradF-value-when-Uneq0}
\nabla F_{ij} &= ((\nabla F \odot U^\star)\mathbf{1}_{m\times m})_{ij} \notag \\
&=  (\nabla F \odot U^\star)_{i:} (\mathbf{1}_{m\times m})_{:j} \notag \\
&= \textstyle\sum_{k=1}^m (\nabla F \odot U^\star)_{ik}.
\end{align}
Thus, on row $i$, for any $j$ such that $U^\star_{ij}\neq0$ we have that the corresponding $\nabla F_{ij}$ is equal to \eqref{eq:gradF-value-when-Uneq0}.
Therefore, all $\nabla F_{ij}$ corresponding to a non-zero $U^\star_{ij}$ on row $i$ are equal.
\end{proof}

\rev{Next, we analyze non-binary stationary points ${U^\star\in\Delta^{m\times m}}$ of \eqref{eq:relaxed-standard-form}.
From the analysis that follows, we will conclude that non-binary stationary points are \emph{strict saddle points}, i.e., critical points
where there is at least one direction along which the curvature is strictly negative.
Note that this definition includes local maxima.
The following proposition gives the necessary and sufficient conditions for identifying a strict saddle point.}

\begin{prop}[\hspace{1sp}\cite{nocedal1999numerical,mokhtari2018escaping}] \label{prop:strict-saddle-point}
Given convex constraints $\mathcal{C}$, a stationary point $x^\star\in\mathcal{C}$ is a \rev{strict} saddle point of the nonlinear problem $\min_{x\in\mathcal{C}}f(x)$ if it satisfies the first-order necessary conditions and there are directions that are undecided to first order, but have negative curvature (i.e., will decrease the objective).
That is, the following conditions must be satisfied
\begin{enumerate}[(i)]
  \item $\nabla f(x^\star)^\top (x-x^\star) \geq 0,\quad\forall\,x\in\mathcal{C}$,
  \item $\exists\,y\in\mathcal{C}$ such that $\nabla f(x^\star)^\top (y-x^\star)=0$ \\ and $(y-x^\star)^\top\nabla^2f(x^\star)(y-x^\star)<0$.
\end{enumerate}
\end{prop}

\rev{To simplify the following analysis, we will use the vectorized form the objective in \eqref{eq:relaxed-standard-form}.}
Let $\mathrm{vecr}:\mathbb{R}^{n\times m}\to\mathbb{R}^{nm}$ be the row-order vectorization operator that stacks matrix rows into a single column.
Let $u\eqdef\mathrm{vecr}(U)$, then the gradient of $F(U)$ in \eqref{eq:relaxed-standard-form} can be expressed as the vectorization
\begin{equation}
\nabla F_\mathrm{v}(u) \eqdef 2(\bar{S}\otimes I)u + 2d(I\otimes P_o + P_d\otimes I) u,
\end{equation}
where $\otimes$ denotes the Kronecker product.

\begin{restatable}{lemma}{nonbinarysaddles}\label{lem:non-binary-stationary-points-are-saddles}
For \rev{$d\ge m+1$}, if a stationary point $U^\star\in\Delta^{m\times m}$ is non-binary, then it is a \rev{strict} saddle point.
\end{restatable}
\begin{proof} 
Assume $U^\star$ has non-binary entries.
Because $\sum_jU^\star_{ij}=1\;\forall_i$, there must be at least two non-zero elements $a$ and $b$ with $0<a\leq b<1$ in a single row.
Without loss of generality, assume that these two non-zero, non-binary elements occur in the first two columns of the first row of $U^\star$.
Let $u^\star\eqdef\mathrm{vecr}(U^\star)$, thus $u^\star=[a,b,\dots]^\top$.
Take $y=[a-\epsilon,b+\epsilon,\dots]^\top$ with $\epsilon>0$ so that $0\leq a-\epsilon<b+\epsilon\leq1$ and let $v\eqdef y-u^\star=[-\epsilon,\epsilon,0,\dots,0]^\top$.

From Lemma~\ref{lem:u-zero-or-gradf-equal}, we know that the entries of $\nabla F_\mathrm{v}(u^\star)$ corresponding to $a$ and $b$ are equal; denote their value as $c$ so that $\nabla F_\mathrm{v}(u^\star)=[c,c,\dots]^\top$.
Therefore
\begin{equation*}
\nabla F_\mathrm{v}(u^\star)^\top v = -c\mkern1mu\epsilon + c\mkern1mu\epsilon = 0,
\end{equation*}
so there exists a $y$ suitable for condition (ii) of Proposition~\ref{prop:strict-saddle-point}.

The curvature of $F_\mathrm{v}$ in the direction of $v$ is given by
\begin{equation*}
v^\top \nabla^2F_\mathrm{v}(u^\star)\, v = 2\,v^\top \left[\bar{S}\otimes I + d(I\otimes P_o + P_d\otimes I)\right] v.
\end{equation*}
Thus, to satisfy condition (ii) of Proposition~\ref{prop:strict-saddle-point}, we must have
\begin{equation}\label{eq:strict-saddle-condition}
    d\,v^\top (I\otimes P_o + P_d\otimes I) v < -v^\top(\bar{S}\otimes I)v,
\end{equation}
We will treat each side separately.
Let $v_1$ denote the first row of $U^\star$.
Then, the right side can be simplified as
\begin{align}
\MoveEqLeft -v^\top(\bar{S}\otimes I)v \notag \\
&=
-\left[
\begin{array}{c}
v_1 \\ \hdashline[2pt/2pt]
0 \\ \hdashline[2pt/2pt]
\vdots \\ \hdashline[2pt/2pt]
0
\end{array}
\right]^\top
\begin{bmatrix}
    \bar{S}_{11} I & \cdots & \bar{S}_{1m} I \\
    \vdots & \ddots & \vdots \\
    \bar{S}_{m1} I & \cdots & \bar{S}_{mm} I
\end{bmatrix}
\left[
\begin{array}{c}
v_1 \\ \hdashline[2pt/2pt]
0 \\ \hdashline[2pt/2pt]
\vdots \\ \hdashline[2pt/2pt]
0
\end{array}
\right] \notag \\
&=
-\bar{S}_{11}\|v_1\|^2.
\end{align}
The left side can be simplified as
\begin{align}
\MoveEqLeft  d\mkern2mu v^\top(I\otimes P_o + P_d\otimes I)v \notag \\
&=
d
\left[
\begin{array}{c}
v_1 \\ \hdashline[2pt/2pt]
0 \\ \hdashline[2pt/2pt]
\vdots \\ \hdashline[2pt/2pt]
0
\end{array}
\right]^\top
\begin{bmatrix}
P_o  & 2I & 2I & \cdots \\
2I & P_o & 2I & \cdots \\
2I & 2I & P_o & \\
\vdots & \vdots & & \ddots
\end{bmatrix}
\left[
\begin{array}{c}
v_1 \\ \hdashline[2pt/2pt]
0 \\ \hdashline[2pt/2pt]
\vdots \\ \hdashline[2pt/2pt]
0
\end{array}
\right] \notag \\
&=
d\mkern4mu v_1^\top P_o \mkern2mu v_1
=
d
\begin{bmatrix} -\epsilon \\ \epsilon \\ 0 \\ \vdots \end{bmatrix}^\top
\begin{bmatrix} \epsilon \\ -\epsilon \\ 0 \\ \vdots \end{bmatrix} \notag \\
&= - d\|v_1\|^2.
\end{align}
Therefore, condition \eqref{eq:strict-saddle-condition} is simplified as
\begin{equation*}
    -d\|v_1\|^2 < -\bar{S}_{11}\|v_1\|^2 \quad\implies\quad d > \bar{S}_{11}.
\end{equation*}
Recall that $|\bar{S}_{ij}|\le1$ and $d\ge m+1$ by assumption.
Hence,
\begin{equation*}
    d \ge m + 1 > 1 \ge |\bar{S}_{11}|,
\end{equation*}
showing that condition (ii) of Proposition~\ref{prop:strict-saddle-point} is satisfied.
\end{proof}

\rev{Lemma~\ref{lem:non-binary-stationary-points-are-saddles} gives us the ability to detect if Algorithm~\ref{alg:mixer} has converged to a strict saddle point (or local maxima).
If it has, then a second-order stationary point may be found by escaping the saddle point.
This is guaranteed using the generic framework proposed in \cite[Alg.~1]{mokhtari2018escaping}, wherein a feasible search direction $v\eqdef y-u^\star$ is found by solving
\begin{gather} \label{eq:descent-dir-qp}
\begin{aligned}
& \underset{y\in\Delta^{m\times m}}{\text{minimize}} & &  q(y;u^\star)\eqdef(y-u^\star)^\top\nabla^2F_\mathrm{v}(u^\star)(y-u^\star) \\
& \text{subject to}
 &&  \nabla F_\mathrm{v}(u^\star)^\top(y-u^\star)=0 \\
\end{aligned}.
\end{gather}
If $q(y;u^\star)<0$ is found, then $u^\star$ is a strict saddle point and $F$ will be decreased by taking a step in the direction of $v$; otherwise $u^\star$ is already at a feasible local minimum.}

\begin{restatable}{lemma}{binarystationarypointsaredistinct}\label{lem:binary-stationary-points-are-distinct}
For \rev{$d\ge m+1$}, if a stationary point $U^\star\in\Delta^{m\times m}$ is binary, then $\phi_\mathrm{dist}(U^\star)=0$, i.e., distinctness is satisfied.
\end{restatable}
\begin{proof} 
Without loss of generality, consider a single view, i.e., $n=1$ and $m=m_1$.
Therefore $(P_d)_{ij} = 2(1-\delta_{ij})$, where $\delta_{ij}=1$ for $i=j$ and $0$ otherwise.

By contradiction, suppose ${U^\star\in\Delta^{m\times m}}$ is a binary stationary point of \eqref{eq:relaxed-standard-form}, but distinctness is violated.
In particular and without loss of generality, assume that the $k$-th column of $U^\star$, denoted $U^\star_{:k}$, is such that ${U^\star_{:k}}^\top\mathbf{1}_m=1+\rho>1$, where $\rho$ is the number of excess entries in violation of distinctness (e.g., if there are $3$ ones in a column then $\rho=2$).

By stationarity \eqref{eq:kkt-stationarity},
\begin{equation}\label{eq:stationarity}
2\bar{S}U^\star + 2d\big( U^\star P_o + P_d U^\star \big) - Y^\star - \lambda^\star\mathbf{1}_m^\top = 0.
\end{equation}
Considering the first row of \eqref{eq:stationarity}, by definition of matrix multiplication we have the terms
\begin{equation*}
2(\bar{S}U^\star)_{1j} = 2\sum^m_{i=1}\bar{S}_{1i}U^\star_{ij} = 2\delta_{kj}\sum^m_{i=1}\bar{S}_{1i},
\end{equation*}
\begin{align*}
2d(U^\star P_o)_{1j} &= 2d\sum^m_{i=1}U_{1i}(P_o)_{ij} = 2d\,U_{1k}(P_o)_{kj} \\
&= 2d(1-\delta_{kj}),
\end{align*}
\begin{align*}
2d(P_dU^\star)_{1j} &= 2d\sum^m_{i=1}(P_d)_{1i}U_{ij} = 4d\sum^m_{i=1}(1-\delta_{1i})U_{ij} \\
&= 4d\sum^m_{i=2}(1-\delta_{1i})U_{ij} = 4d\sum^m_{i=2}U_{ij} \\
&= 4d\rho\delta_{kj},
\end{align*}
so that the $j$-th entry in the first row of \eqref{eq:stationarity} can be written
\begin{equation}\label{eq:stationarity-row}
2\delta_{kj}\sum^m_{i=1}\bar{S}_{1i} + 2d(1-\delta_{kj}) + 4d\rho\delta_{kj} = Y^\star_{1j} + \lambda^\star_1.
\end{equation}
Since $U^\star_{1k}=1$ by assumption, $Y^\star_{1k}=0$ by complementarity~\eqref{eq:kkt-complementarity}.
Therefore, solving \eqref{eq:stationarity-row} for $\lambda^\star_1$ when $j=k$ yields
\begin{equation*}
\lambda^\star_1 = 2\sum^m_{i=1}\bar{S}_{1i} + 4d\rho.
\end{equation*}
Then, when $j\neq k$, we can use $\lambda^\star_1$ to solve for $Y^\star_{1j}$ as
\begin{equation*}
Y^\star_{1j} = 2d - 4d\rho - 2\sum^m_{i=1}\bar{S}_{1i}.
\end{equation*}
By dual feasibility \eqref{eq:kkt-dual-feasibility}, $Y^\star_{1j}\ge0$ so
\begin{equation*}
d(1-2\rho) \ge \sum^m_{i=1}\bar{S}_{1i} \ge -m
\end{equation*}
would guarantee dual feasibility, where the lower bound $-m$ is due to $|\bar{S}_{ij}|\le1$.
By assumption, $d\ge m+1$, and assuming the worst case where all entries violate distinctness (i.e., ${\rho=m-1}$) gives the tightest bounds
\begin{equation*}
d(1-2\rho) \ge (m+1)(1-2\rho) \ge 2(1-m^2) \ge -m,
\end{equation*}
implying that $-2m^2+m+2\ge0$ for dual feasibility \eqref{eq:kkt-dual-feasibility} to hold, which is a contradiction for $m>1$.
\end{proof}

\rev{Given the characterization of non-binary and binary stationary points in Lemma~\ref{lem:non-binary-stationary-points-are-saddles} and Lemma~\ref{lem:binary-stationary-points-are-distinct}, we now present our main result concerning \eqref{eq:relaxed} (which was restated as \eqref{eq:relaxed-standard-form}).}

\feasibilityoflocalminima*
\begin{proof}
By direct consequence of Lemma~\ref{lem:non-binary-stationary-points-are-saddles}, when $d\ge m+1$, non-binary solutions are strict saddles.
In this case, by solving \eqref{eq:descent-dir-qp} strict saddles can be escaped, ensuring convergence to a second-order stationary point $U^\star$, i.e., a (local) minima~\cite[Theorem 4]{mokhtari2018escaping}.
By the contrapositive of Lemma~\ref{lem:non-binary-stationary-points-are-saddles}, it follows that $U^\star$ is binary, and therefore must be distinct due to Lemma~\ref{lem:binary-stationary-points-are-distinct}.
Since $U^\star$ is cycle consistent by construction~\cite{zhou2015multi}, the proof is complete.
\end{proof}

\section{Update Rule Analysis}\label{apdx:update-rule}

\rev{We must increment the non-negative parameter $d$ until $U\in\Delta^{m\times m}$ is binary and satisfies all the constraints of \eqref{eq:penalty-form}.
Our strategy for incrementing $d$ is motivated by the desire to quickly converge to a feasible solution.
Focusing on the elements of $U$ that contribute to the violation of distinctness and orthogonality allows us to do so.
In what follows we refer to these two constraints together simply as the penalty $\Phi(U) \eqdef \phi_\mathrm{orth}(U) + \phi_\mathrm{dist}(U)$, with $\nabla\Phi=UP_o+P_dU$.}

\rev{Observe that $\nabla\Phi$ is always non-negative and that non-zero entries $\nabla\Phi_{ij}$ indicate the entries of $U$ that if increased would incur more penalty---these are the \emph{potentially} problematic entries of $U$.
Note that all the entries of $U$ are in $[0,1]$ and only the non-zero entries could contribute to distinctness or orthogonality violations.
Thus, to precisely identify the problematic entries, we find entries satisfying $\nabla\Phi_{ij}>0,U_{ij}>0$.
For each of those problematic entries, we solve for the value of $d$ that would cause $\nabla F_{ij}\geq0$ so that the step $-\nabla F$ causes $U_{ij}$ to diminish.
The set of values that have this property for problematic entries is (i.e., set $\nabla F=0$ and solve for $d$)
\begin{equation}
    \mathcal{D} = \left\{ -\frac{[\mathbf{1}-2S]}{[\nabla\Phi]} : [\nabla\Phi]>0,[U]>0 \right\},
\end{equation}
where the notation $[\cdot]$ indicates an element-wise operation.
The number of entries diminished to zero is controlled by taking the maximum, median, or minimum of this set.}

\rev{We analyze ten different update rules for $d$ in Table~\ref{tbl:d_update_rule}.
The value of $d$ chosen at each outer iteration is given by the sequence under the ``Update Rule'' column.
Our study indicates that the best tradeoff between a low number of outer iterations and high accuracy can be achieved by first selecting $\mathrm{med}\,\mathcal{D}$ and then doubling the value of $d$ each subsequent outer iteration.
This method takes a principled approach to selecting the first penalty value based on the problem data and initial guess, followed by doubling the penalty weight to quickly lead to convergence to a feasible, binary solution.
Note that initializing $d$ too large leads to abysmal accuracy.}

\begin{table}[h]
\footnotesize
\centering
\caption{
\rev{Analysis of update rules for $d$ (Line 10) in Algorithm~\ref{alg:mixer}.
Ten rules are evaluated on the ten datasets of Table~\ref{tbl:benchmark_results} and Table~\ref{tbl:carfusion_results}.
We compare the average number of outer iterations necessary until convergence and the average solution accuracy. %
$\mathrm{F}_1$ score is normalized according to the maximum $\mathrm{F}_1$ across methods within a given dataset.
The strategy that achieves the highest accuracy begins with $d=0$ and slowly increments $d$ by $0.1$ each outer iteration, but requires more than $50$ outer iterations on average.
A better tradeoff between high accuracy and low iteration count can be found by first selecting $\mathrm{med}\,\mathcal{D}$ and then doubling the value of $d$ each subsequent outer iteration.}
}
\setlength{\tabcolsep}{0pt}
\ra{1.0}
\sisetup{
    detect-weight=true,
    separate-uncertainty=true,
    }
\begin{tabular}{@{}l S S@{}}
\toprule
Update Rule & {Num. Outer Iters.} & {Normalized $\mathrm{F}_1$} \\
\toprule
$d_i=0,\max\mathcal{D},2d_{i-1},\dots$ & 2.9\pm1.2 & 0.927\pm0.05 \\
$d_i=0,\mathrm{med}\,\mathcal{D},2d_{i-1},\dots$ & 3.7\pm0.5 & 0.938\pm0.05 \\
$d_i=0,\min\mathcal{D},2d_{i-1},\dots$ & 6.1\pm1.1 & 0.942\pm0.04 \\
$d_i=\max\mathcal{D},2d_{i-1},\dots$ & 2.5\pm1.3 & 0.917\pm0.07 \\
\boldmath $d_i=\mathrm{med}\,\mathcal{D},2d_{i-1},\dots$ & \bfseries 4.8\pm0.9 & \bfseries 0.980\pm0.03 \\
$d_i=\min\mathcal{D},2d_{i-1},\dots$ & 15.6\pm1.2 & 0.936\pm0.05 \\
$d_i=0,0.1,2d_{i-1},\dots$ & 8\pm0.8 & 0.950\pm0.04 \\
$d_i=0,0.1,0.2,\dots$ & 52.5\pm28.4 & 0.988\pm0.03 \\
$d_i=1,2d_{i-1},\dots$ & 3.6\pm2.2 & 0.944\pm0.05 \\
$d_i=10,2d_{i-1},\dots$ & \num{1 (0)} & 0.238\pm0.19 \\
\bottomrule
\end{tabular}
\label{tbl:d_update_rule}
\end{table}

\section{Timing Analysis}\label{apdx:timing}

\begin{figure*}[t]
    \centering
    \begin{subfigure}[b]{1\textwidth}
        \includegraphics[trim=0mm 0mm 0mm 0mm, clip, width=1\columnwidth]{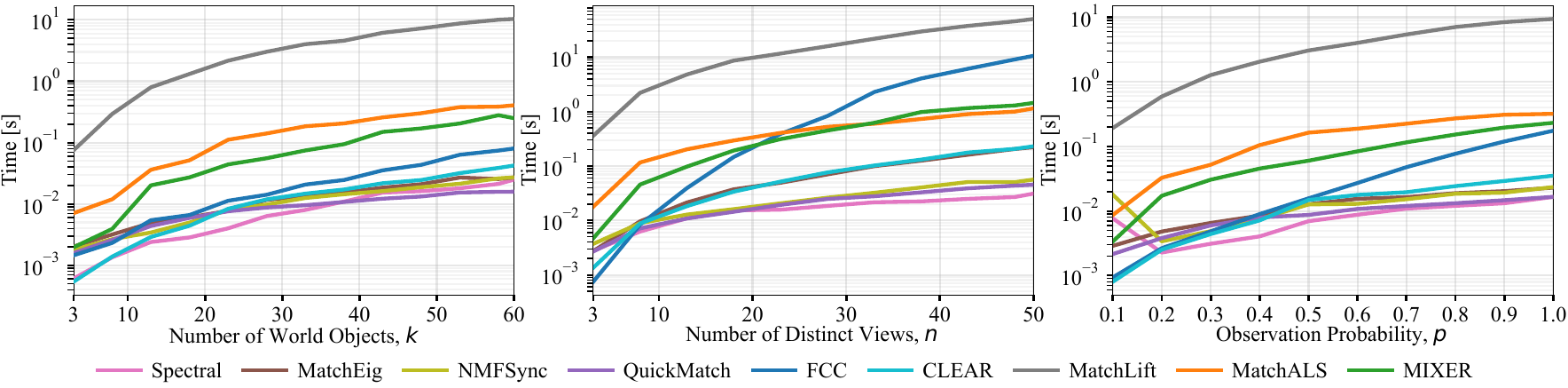}
    \end{subfigure}
    \caption{
    Runtime comparison when operating on synthetic data with various problem sizes.
    Problem size is a function of the number of world objects $k$, the number of views $n$, and the probability $p$ of a given view observing all $k$ objects.
    Nominal values for these parameters are $k=30$, $n=10$, and $p=0.5$.
    Each subplot is generated by varying one of the parameters while holding the other two constant.
    In general, MIXER achieves the highest accuracy (see Section~\ref{sec:experiments}) while being faster than its closest competitors MatchLift and MatchALS (see Table~\ref{tbl:relaxations}).
    Although other algorithms are faster than MIXER, they frequently fail to return accurate solutions in realistic settings with a moderate number of world objects and low observation probability.
    }
    \label{fig:synthetic_timing}
    \vskip-1em
\end{figure*}

\rev{We compare the runtime of Algorithm~\ref{alg:mixer} to existing algorithms.
Runtime complexity is driven by the problem size $m$, which is a function of the number of world objects $k$, the number of views $n$, and the probability $p$ of a given view observing all $k$ objects.
Therefore, we generating synthetic data with nominal values of $k=30$, $n=10$, and $p=0.5$ and produce three plots in Fig.~\ref{fig:synthetic_timing} corresponding to varying one parameter while holding the other two at their nominal values.
The nominal values are chosen for consistency with the synthetic experiments in Section~\ref{sec:synthexpts}, with results in Fig.~\ref{fig:synthetic_results_f1} and Fig.~\ref{fig:synthetic_results_optgap}.
For all three plots in Fig.~\ref{fig:synthetic_timing}, we use a mismatch of $25\%$ and the runtime at each parameter setting is averaged over $10$ trials.}

\rev{Fig.~\ref{fig:synthetic_timing} shows that in general the fastest algorithm is Spectral while the slowest is MatchLift.
MIXER is faster than its most similar competitors, MatchLift and MatchALS (see Table~\ref{tbl:relaxations}).
Indeed, these competitors are the closest in terms of accuracy (e.g., Table~\ref{tbl:carfusion_results}), but are considerably slower than MIXER.
The remaining algorithms (Spectral, MatchEig, NMFSync, CLEAR) perform well in terms of runtime and obtain similar execution speeds to each other.
Although these algorithms are faster than MIXER, previous results (Figs.~\ref{fig:synthetic_results_f1}~and~\ref{fig:synthetic_results_optgap}, Tables~\ref{tbl:benchmark_results}~and~\ref{tbl:carfusion_results}) indicate that they are not able to achieve the high accuracy that MIXER does.
The FCC algorithm varies in runtime the most, with the largest change in runtime occurring when a large number of distinct views (i.e., large $n$) exists in the dataset.
In realistic scenarios with a moderate amount of world objects (e.g., using local submaps to reduce the number of objects needed to be processed), there are frequently low observation probabilities as robot sensors are noisy and field-of-view-limited so that only a subset of the world objects are seen in each view.
In this scenario (e.g., Section~\ref{sec:carexpt}), our unoptimized MATLAB implementation of MIXER achieves the best accuracy at runtimes on the order of a few hundred milliseconds.}

\balance

\end{document}